\documentclass[11pt]{article}

\usepackage[final]{acl}

\usepackage{times}
\usepackage{latexsym}

\usepackage[T1]{fontenc}

\usepackage[utf8]{inputenc}

\usepackage{microtype}

\usepackage{inconsolata}

\usepackage{graphicx}

\usepackage[table]{xcolor}
\usepackage{hyperref}
\usepackage{url}
\usepackage{amssymb}
\usepackage{subcaption}
\usepackage{amsmath}  
\usepackage{amssymb}  
\usepackage{bm}       
\usepackage{booktabs}      
\usepackage{multirow}      
\usepackage{array}         
\usepackage{amsmath}       
\usepackage{graphicx}      
\usepackage{caption}       

\title{Polynomial Expansion Rank Adaptation: Enhancing Low-Rank Fine-Tuning with High-Order Interactions}

\author{Wenhao Zhang$^{1}$ \and Lin Mu$^{1}$\footnotemark \and  Li Ni$^{1}$  \and Peiquan Jin$^{2}$ \and Yiwen Zhang$^{1}$ \\
        $^{1}$ Anhui University,\\
        $^{2}$ University of Science and Technology of China,\\
       \small{ \texttt{\{mulin, nili, zhangyiwen\}@ahu.edu.cn}   \texttt{zhangwenhao@stu.ahu.edu.cn}} \\
        \small{\texttt{jpq@ustc.edu.cn}}
}

\begin{document}
\maketitle
\begin{abstract}
Low-rank adaptation (LoRA) is a widely used strategy for efficient fine-tuning of large language models (LLMs), but its strictly linear structure fundamentally limits expressive capacity. The bilinear formulation of weight updates captures only first-order dependencies between low-rank factors, restricting the modeling of nonlinear and higher-order parameter interactions.
In this paper, we propose \textbf{P}olynomial \textbf{E}xpansion \textbf{R}ank \textbf{A}daptation (\textbf{PERA}), a novel method that introduces structured polynomial expansion directly into the low-rank factor space.
By expanding each low-rank factor to synthesize high-order interaction terms before composition, PERA transforms the adaptation space into a polynomial manifold capable of modeling richer nonlinear coupling without increasing rank or inference cost.
We provide theoretical analysis demonstrating that PERA offers enhanced expressive capacity and more effective feature utilization compare to existing linear adaptation approaches.
Empirically, PERA consistently outperforms state-of-the-art methods across diverse benchmarks. Notably, our experiments show that incorporating high-order nonlinear components—particularly square terms—is crucial for enhancing expressive capacity and maintaining strong and robust performance under various rank settings. Our code is available at \url{https://github.com/zhangwenhao6/PERA} 


\end{abstract}
\renewcommand{\thefootnote}{\fnsymbol{footnote}}
\footnotetext[1]{Corresponding author}

\section{Introduction}
Large language models~\citep{brown2020language, ouyang2022training, touvron2023llama} (LLMs) are typically trained on large-scale general-domain datasets using the next-token prediction objective~\citep{brown2020language}, and exhibit remarkable generalization capabilities across a wide range of natural language processing tasks~\citep{thirunavukarasu2023large,ruiz2023dreambooth}.
However, as model sizes continue to grow and downstream tasks become increasingly diverse, the computational and memory costs associated with full fine-tuning have risen dramatically~\citep{touvron2023llama2,mistral7b}, thereby constraining the widespread deployment of LLMs.
\begin{figure}[t]
    \centering
    \includegraphics[width=0.9\linewidth]{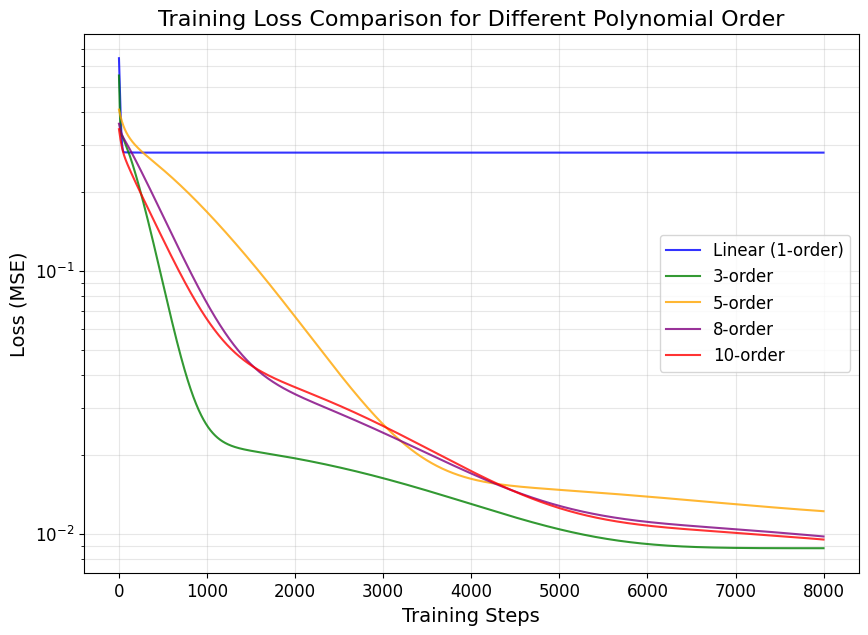}
    \caption{Comparison of Training Losses between First-Order and High-Order Terms.}
    \label{fig:fig1}
\end{figure}

To address this challenge, parameter-efficient fine-tuning (PEFT)~\citep{houlsby2019parameter,liu2021p,lialin2023scaling} has emerged as an effective paradigm that adapts LLMs by introducing a small number of task-specific parameters while keeping the backbone weights frozen. Among various PEFT methods, low-rank adaptation (LoRA)~\cite{hu2022lora} achieves strong efficiency by restricting weight updates to a low-rank subspace. Nevertheless, LoRA’s strictly bilinear update form $\Delta W=BA$ inherently restricts the expressivity of the adaptation, as it captures only first-order linear dependencies between the low-rank factors, restricting its ability to model multi-dimensional or high-order dependencies ~\citep{liu2023chipnemo,jiang2024mora,zhuang2024time}.
To partially mitigate this, HiRA~\citep{huang2025hira} introduces Hadamard modulation with pretrained weights, enriching multi-dimensional representation. However, its update mechanism remains fundamentally linear with respect to trainable parameters and depends on external weight coupling, thus offering limited ability to model intrinsic high-order relations.

From the perspective of function approximation, there exists a fundamental difference in expressive capacity within function space between a first-order linear function $f(x)=c+c_1x$ and a polynomial function with higher-order terms $f(x)=c+c_1x+c_2x^2+\dots+c_nx^n$, as illustrated in Fig.~\ref{fig:fig1}. This directly leads to significant differences in fitting accuracy, convergence speed, and training loss between the two. If the traditional LoRA mapping is viewed as a first-order linear approximation of weight updates, its limitation in expressive capacity becomes immediately evident.

Motivated by these observation, we investigate whether the expressive capacity of low-rank adaptation can be enhanced by introducing explicit high-order feature interactions and structured nonlinearity directly within the low-rank factor space—without increasing rank or depending on external modulation. Drawing inspiration from classical polynomial feature expansion~\citep{polynomial_expansion2, polynomial_expansion1, polynomial_expansion3}, which systematically constructs higher-order feature terms from base representations, we propose \textbf{P}olynomial \textbf{E}xpansion \textbf{R}ank \textbf{A}daptation (\textbf{PERA}).

Unlike traditional methods that perform polynomial expansion in the feature space, PERA applies a structured polynomial mapping directly within the parameter space of low-rank factors. Specifically, before composition, each low-rank matrix is expanded along its column dimension to generate higher-order parameter interaction terms. This operation constructs a polynomial manifold in the adaptation space, enabling $\Delta W$ to capture richer nonlinear coupling relationships among adaptation directions. Empirical analyses demonstrate faster convergence and reduced training loss. Furthermore, our theoretical results indicate that this formulation significantly enhances both the expressive capacity of adaptation and the efficiency of feature utilization. Notably, we implement higher-order interaction terms via matrix concatenation rather than sequential matrix addition. As a result, no additional inference overhead is introduced, while preserving the modular efficiency of LoRA.

Our contributions can be summarized as follows:
\begin{itemize}
    \item We propose PERA that introduces polynomial expansion within the parameter space of low-rank factors.
    PERA explicitly models high-order interactions and structured nonlinearity among adaptation directions, enhancing representational expressivity without increasing rank or parameter cost.
    
    \item We theoretically show that parameter-space polynomial expansion improves both the expressive capacity and feature utilization efficiency of low-rank adaptation, offering a principled explanation for its stronger representation power.
    
    \item We empirically demonstrate that, compared to other PEFT methods, our approach maintains a computational and memory footprint close to stan-dard LoRA and achieves lower training loss. 
\end{itemize}

\begin{figure*}[t]
    \label{fig1}
    \centering
    \includegraphics[width=0.95\linewidth]{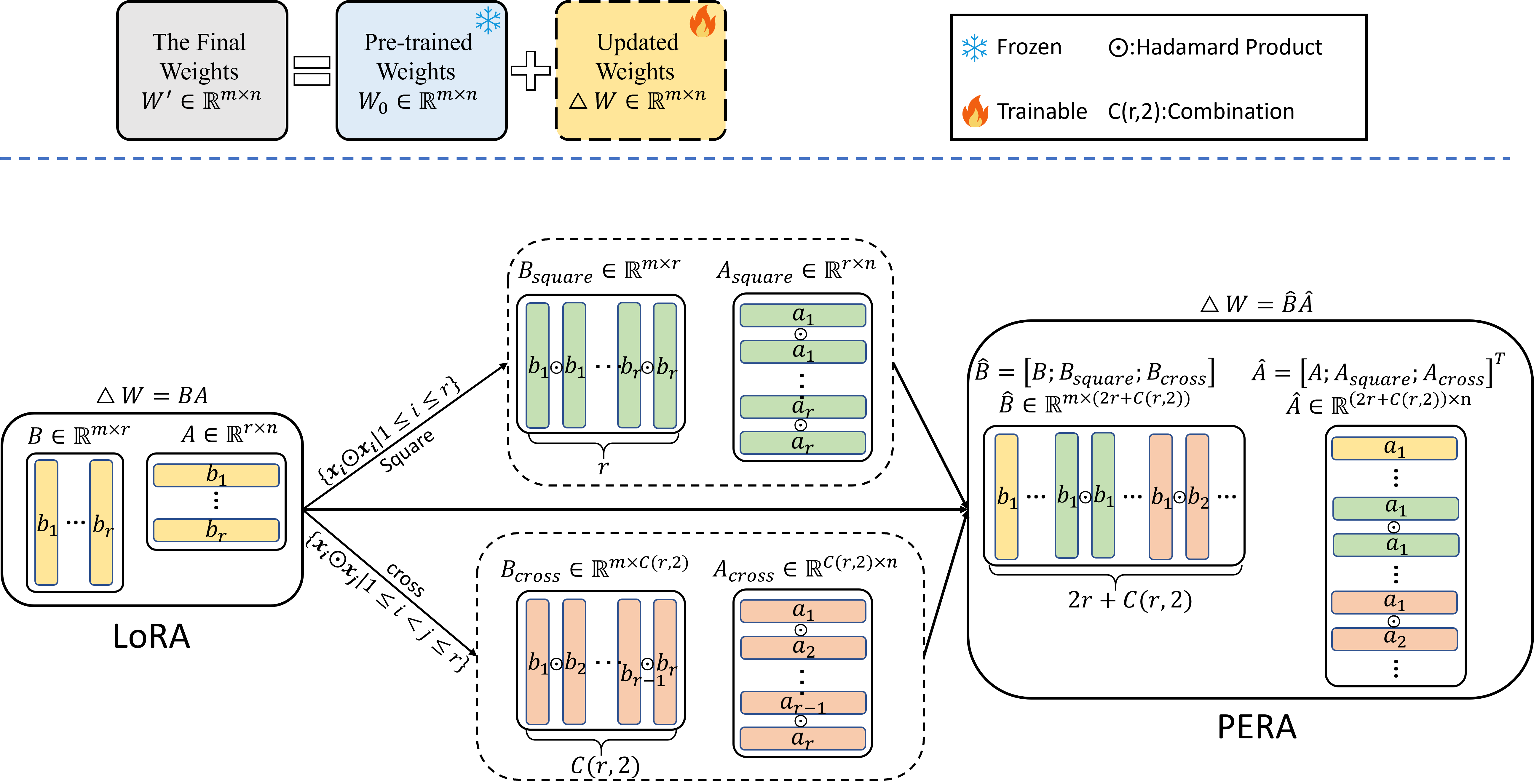}
    \caption{The architecture comparison between LoRA and PERA. By applying the polynomial expansion technique, PERA synthesizes new high-order feature terms from the original $r$ feature terms in the low-rank matrices, including \textbf{$r$ square feature terms} and \textbf{$C(r,2)$ crossed feature terms.}}
    \label{fig:fig2}
\end{figure*}

\section{Related Work}
\textbf{Parameter Efficient Fine-Tuning.} 
Parameter-efficient fine-tuning (PEFT) aims to substantially reduce the number of trainable parameters while maintaining model performance and efficiency. To this end, researchers have explored a variety of approaches. For instance, Adapter~\citep{hu2023llm-adapters} inserted lightweight feed-forward network modules into Transformer~\cite{Attention} layers and updates only these newly introduced parameters for downstream adaptation. Prefix-tuning~\citep{li2021prefix} learned continuous prefix vectors that encode task-specific information and steer the model toward generating task-relevant outputs. Prompt-tuning~\citep{lester2021power} further simplified this paradigm by optimizing only the embedding representations of input prompts, thereby effectively activating the model's capabilities on downstream tasks. Despite their strong performance, these methods typically introduce additional computational overhead during inference.

\textbf{Low-Rank Adaptation (LoRA).} LoRA~\citep{hu2022lora} leverages the observation that weight updates in pretrained models have low intrinsic rank. It introduces two trainable low-rank matrices, $A$ and $B$, into linear layers to approximate task-specific updates without altering the architecture or adding inference latency. Given a pretrained weight $W_{0}$, the adapted weight is defined as:
\begin{equation}
    \label{eq1}
    W' = W_{0} + \Delta W = W_{0} + \frac{\alpha}{r}BA,
\end{equation}
where $\alpha$ is a scaling factor and $r$ denotes the rank, with $r\ll \min\{m,n\}$. During fine-tuning, the pretrained weights $W_{0}$ remain frozen, and the model is adapted by optimizing the low-rank matrices $A$ and $B$. 
Building upon LoRA, DoRA\citep{liu2024dora} decomposed the weight updates into magnitude and directional components, applying normalization and scaling to each singular column of $W'$. Similarly, DeLoRA~\citep{bini2025delora} performed normalization within the internal $r$-dimensional space of each low-rank matrix. 
To learn high-rank weight updates, MoRA\citep{jiang2024mora} compressed, transformed, and decompressed the input to enable high-rank adaptation, while HiRA\citep{huang2025hira} increased the maximum achievable rank by applying the Hadamard product between the updated weights $\Delta W$ and the original weights $W$.

Unlike HiRA and LoRA, PERA employs polynomial expansion to generate explicit high-order feature terms from low-rank factors. This design enhances nonlinear expressivity and cross-dimensional interactions while maintaining LoRA’s efficiency and zero inference overhead.


\section{Methodology}
\label{3}


\subsection{Preliminaries: Polynomial Expansion}
\label{3.1}
Polynomial expansion~\citep{polynomial_expansion2, polynomial_expansion1, polynomial_expansion3} is a commonly used technique in feature engineering, primarily employed to generate synthetic features from original representations when feature dimensionality is limited. 

Formally, consider a matrix $B\in \mathbb{R}^{m\times r}$ whose column vectors are denoted as  $B = [ \mathbf{b}_{1}, \mathbf{b}_{2}, \cdots, \mathbf{b}_{r} ]$, where each $\mathbf{b}_{i}\in \mathbb{R}^{m}$. We define a $k$-th order polynomial expansion operation as $\text{Poly}^{\text{k}}(B)$, which constructs all possible $k$-order polynomial combinations among the columns of $B$.
For computational efficiency, we typically adopt the $2$-order polynomial expansion, formula as: 
\begin{equation}
\label{eq2}
\text{Poly}^2(B)=
\left[
\mathbf{b}_i,
\mathbf{b}_i \odot \mathbf{b}_j
\;\middle|\; 1\le i\le j \le r
\right]
\end{equation}

where $\odot$ denotes element-wise multiplication (Hadamard product). The expanded feature set can be written as $\hat{B} = [{B;B_{square};B_{cross}}]$, consisting of $r$ original features, $r$ square features and $C(r,2)$ pairwise cross features:
\begin{equation*}
\begin{aligned}
B &= \left\{ \mathbf{b}_i \,\middle|\, 1 \le i \le r \right\}, \\
B_{square} &= \left\{ \mathbf{b}_i \odot \mathbf{b}_j \,\middle|\, 1 \le i =j \le r \right\}, \\
B_{cross} &= \left\{\, \mathbf{b}_i \odot \mathbf{b}_j \;\middle|\; 1 \le i < j \le r \,\right\}
\end{aligned}
\end{equation*}
This expansion maps the original low-rank space \( \mathbb{R}^{r} \) to a higher dimensional space \( \mathbb{R}^{2r+C(r,2) }\).

To further enhance representation capacity, we extend this formulation by introducing a Hadamard-based polynomial expansion. Similarly, consider a matrix $A\in\mathbb{R}^{r\times n}$ whose row vectors are represented as $A = [ \mathbf{a}_{1}, \mathbf{a}_{2}, \cdots, \mathbf{a}_{r} ]^T$, where each $\mathbf{a}_{i} \in \mathbb{R}^{n}$. We define the Hadamard-based polynomial expansion as:
\begin{multline}
\label{eq3}
    \text{Poly}^{2}_{\text{H}}(A) = \\ 
    [ \mathbf{a}_i, h_{ij} (\mathbf{a}_i \odot \mathbf{a}_j) \mid 1 \le i \leq j \le r ]^T.
\end{multline}
Accordingly, this expansion can be defined as $\hat{A}= [{A;A_{square};A_{cross}}]^T$: 
\begin{equation*}
\begin{aligned}
A &= \left\{ \mathbf{a}_i \,\middle|\, 1 \le i \le r \right\}, \\
A_{square} &= \left\{ h_{ij}(\mathbf{a}_i \odot \mathbf{a}_j) \,\middle|\, 1 \le i=j \le r \right\}, \\
A_{cross} &= \left\{h_{ij}(\mathbf{a}_i \odot \mathbf{a}_j) \;\middle|\; 1 \le i < j \le r \,\right\} 
\end{aligned}
\end{equation*}
here, $\mathbf{h} =\{h_{ij} \mid 1 \le i \leq j \le r\}$ is learnable coefficients initialized to zero for stability. This initialization strategy ensures smooth optimization while allowing the model to gradually incorporate nonlinear contributions from high-order features.

\subsection{PERA Architecture}
\label{3.2}
We propose PERA, a simple yet effective PEFT method that introduces polynomial expansion into the low-rank adaptation space. Inspired by polynomial techniques in feature engineering, PERA enhances the expressive power of weight updates by enabling explicit high-order interactions among low-rank factors—without increasing the nominal rank. Following the LoRA framework, we decompose the weight update into two trainable low-rank matrices $B \in \mathbb{R}^{m\times r}$ and $A\in \mathbb{R}^{r\times n}$, initialized with a Gaussian distribution and zero values, respectively. Then, we apply structured polynomial mappings to these matrices: For $B$, we perform a standard polynomial expansion $\text{Poly}^{2}(B)$ (Eq.~\ref{eq2}). For $A$,  we employ the Hadamard-based polynomial expansion $\text{Poly}^{2}_{\text{H}}(A)$ (Eq.~\ref{eq3}) to ensure stability during optimization. The resulting parameter update is defined as:
\begin{equation}
    \Delta W = \hat{B}\hat{A} = \text{Poly}^{2}(B)\text{Poly}^{2}_{\text{H}}(A),
    \label{eq4}
\end{equation}
where, $\text{Poly}^{2}(B) \in \mathbb{R}^{m\times (2r+C(r,2))}$ denotes the expanded matrix obtained by applying polynomial expansion to the low-rank matrix $B$, and $\text{Poly}^{2}_{\text{H}}(A) \in \mathbb{R}^{ (2r+C(r,2))\times n}$ represents the matrix produced by applying the Hadamard-based polynomial expansion to the low-rank matrix $A$.

We integrate the frozen pretrained weight $W_{0}$ with the adapted hidden representation. Specifically, for a linear layer with forward computation $h = W_{0}x$, where $x \in \mathbb{R}^{n}$ denotes the input representation, PERA modifies its forward computation as follows:
\begin{equation}
\label{eq5}
\begin{aligned}
    h &= W' x \\
      &= W_{0} x +\text{Poly}^{2}(B)\, \text{Poly}^{2}_{\text{H}}(A)\, x.
\end{aligned}
\end{equation}
Notably, during backpropagation, gradients are computed only with respect to the low-rank matrices $A$ and $B$, along with the Hadamard vector $\mathbf{h}$. As a result, compared to the LoRA, PERA does not introduce significant additional parameter overhead while substantially improving expressive capacity.

\subsection{Analysis of PERA}
\label{3.3}
\subsubsection{Rank Analysis}
\label{3.3.1}
LoRA and its variants construct the weight update $\Delta W$ as the product of two low-rank matrices, which inherently limits the maximum achievable rank of the updated weight. Let the pretrained weight matrix be $W_{0}\in \mathbb{R}^{m\times n}$ with rank $r_{0}$, where $r_{0} \le min(m,n)$. According to Eq.~\ref{eq1} and basic properties of matrix rank, the rank of the updated weight $W' = W_0 + \Delta W$ satisfies: 
\begin{equation}
\label{eq6}
    \mathrm{Rank}(W') \leq r_{0}+r.
\end{equation}
This implies that the low-rank structure of $\Delta W$ restricts the rank growth of $W'$, potentially limiting the expressiveness of the adapted model.
In contrast, PERA applies polynomial expansion to the low-rank factors, resulting in an expanded factor dimension of $2r+C(r,2)$(Eq.~\ref{eq5}). Consequently, the rank of the adapted weight in PERA is upper-bounded by:
\begin{equation}
\label{eq7}
    \mathrm{Rank}(W') \leq r_{0}+ (2r+C(r,2)).
\end{equation}
Compared with Eq.~\ref{eq6}, PERA substantially increases the theoretical upper bound on the rank of the adapted weight matrix, thereby enlarging the space of feasible updates and enabling more expressive adaptation under the PEFT framework. A detailed proof is provided in Appendix~\ref{Appendix: A.1}.

\subsubsection{Feature Utilization Analysis}
\label{3.3.2}
Another key difference between PERA and LoRA lies in the expressive form of the weight update. In LoRA, the update matrix is constructed as the product of two low-rank matrices $B\in \mathbb{R}^{m\times r}$ and $A\in \mathbb{R}^{r\times n}$. By representing $B$ and $A$ in terms of their rank-one components: $B = [\mathbf{b}_{1}, \mathbf{b}_{2}, \cdots, \mathbf{b}_{r}]$, the LoRA update can be expressed as:
\begin{equation}
    \label{eq8}
    \Delta W = BA = \sum_{i=1}^{r}\mathbf{b}_{i}\mathbf{a}_{i}^T,
\end{equation}
which is a linear combination of rank-one matrices.

In contrast, PERA enriches the low-rank update by introducing polynomial feature expansions on the low-rank factors. Specifically, PERA incorporates both square feature terms and pairwise cross terms through polynomial expansion of $B$ and a Hadamard-based polynomial expansion of $A$. As a result, the update matrix in PERA can be expressed as:
\begin{equation}
\label{eq9}
\begin{aligned}
\Delta W 
& = \text{Poly}^{2}(B)\text{Poly}^{2}_{\text{H}}(A) \\
&= \sum_{i=1}^{r} \mathbf{b}_{i} \mathbf{a}_{i}^{T} \\
&\quad + \sum_{1\le i =j}^{r} h_{ij}(\mathbf{b}_{i}\odot \mathbf{b}_{j}) (\mathbf{a}_{i}^{T}\odot \mathbf{a}_{j}^{T}) \\
&\quad + \sum_{1 \le i < j}^r h_{ij}(\mathbf{b}_{i} \odot \mathbf{b}_{j})(\mathbf{a}_{i}^T \odot \mathbf{a}_{j}^T).
\end{aligned}
\end{equation}
Comparing Eq.~\ref{eq8} and Eq.~\ref{eq9}, we observe that PERA explicitly augments the linear low-rank update with square and cross feature terms, introducing structured high-order nonlinear components into $\Delta W$. This enriched formulation enables more diverse feature utilization and substantially enhances the expressive capacity of the adaptation, while preserving the factorized structure of low-rank updates. A detailed theoretical analysis is provided in Appendix~\ref{Appendix: A.2}.


\begin{table*}[htbp]
\centering
\small
\resizebox{\textwidth}{!}{
\begin{tabular}{l l c c c c c c c c c c}
\toprule
\textbf{Model} & \textbf{Method} & \textbf{\#Params (\%)} & \textbf{BoolQ} & \textbf{PIQA} & \textbf{SIQA} & \textbf{ARC-c} & \textbf{ARC-e} & \textbf{OBQA} & \textbf{HellaS} & \textbf{WinoG} & \textbf{Avg.} \\
\midrule
ChatGPT & - & - & 73.10 & 85.40 & 68.50 & 79.90 & 89.80 & 74.80 & 78.50 & 66.10 & 77.01 \\
\midrule
\multirow{9}{*}{Llama-2-7B} 
& Prompt Tuning & 0.0012 & 55.93 & 12.35 & 30.50 & 6.06 & 8.63 & 9.40 & 6.91 & 40.57 & 21.29 \\
& P-Tuning & 0.7428 & 58.75 & 36.02 & 0.20 & 0.17 & 1.98 & 0.80 & 0.01 & 0.00 & 12.24 \\
& LoRA ($r=32$) & 0.8256 & 69.80 & 79.90 & 79.50 & 64.70 & 79.80 & 81.00 & 83.60 & 82.60 & 77.61 \\
& DoRA ($r=32$) & 0.8256 & 71.80 & 83.70 & 76.00 & 68.20 & 83.70 & 82.40 & \textbf{89.10} & 82.60 & 79.69 \\
& MoRA ($r=32$) & 0.8241 & 72.17 & 80.79 & 79.53 & 71.42 & 85.31 & 81.20 & 29.09 & 80.19 & 72.46 \\
& HiRA ($r=32$) & 0.8256 & 71.22 & 83.35 & 79.53 & 73.81 & 86.74 & 84.60 & 88.12 & 83.98 & 81.42 \\
& \textbf{PERA} ($r=16$) & 0.4148 & \underline{71.83} & \textbf{85.31} & \underline{80.35} & \textbf{73.55} & \textbf{88.55} & \textbf{85.40} & 88.66 & \textbf{87.13} & \textbf{82.61} \\
& \textbf{PERA} ($r=32$) & 0.8268 & \textbf{73.70} & \underline{84.39} & 79.53 & \underline{73.29} & 87.79 & \underline{83.20} & \underline{88.98} & \underline{86.27} & \underline{82.14} \\

\midrule
\multirow{10}{*}{Llama-3-8B} 
& Prompt Tuning & 0.0010 & 56.85 & 45.05 & 36.13 & 31.57 & 32.74 & 29.20 & 14.01 & 50.12 & 36.96 \\
& P-Tuning & 0.6240 & 59.97 & 11.64 & 8.19 & 7.42 & 8.63 & 9.60 & 1.77 & 37.65 & 18.11 \\
& LoRA ($r=16$) & 0.3513 & 72.30 & 86.70 & 79.30 & 75.70 & 87.70 & 82.80 & 93.50 & 84.80 & 82.80 \\
& LoRA ($r=32$) & 0.7002 & 70.80 & 85.20 & 79.90 & 71.20 & 84.20 & 79.00 & 91.70 & 84.30 & 80.79 \\
& DoRA ($r=32$) & 0.7002 & 74.60 & 89.30 & 79.90 & 80.40 & 90.50 & 85.80 & 95.50 & 85.60 & 85.20 \\
& MoRA ($r=32$) & 0.6997 & 74.28 & 87.43 & 80.71 & 79.61 & 91.16 & 85.60 & 43.53 & 86.74 & 78.63 \\
& HiRA ($r=16$) & 0.3513 & 73.85 & 89.12 & 81.06 & 82.59 & 93.06 & 87.40 & 94.85 & 86.74 & 86.08 \\
& HiRA ($r=32$) & 0.7002 & 75.40 & 89.70 & 81.15 & 82.90 & 93.27 & 88.32 & 95.36 & 87.70 & 86.72 \\
& \textbf{PERA} ($r=16$) & 0.3515 & \textbf{76.15} & 88.57 & \textbf{83.16} & \textbf{83.62} & \textbf{93.73} & \textbf{88.60} & \underline{95.94} & \textbf{89.27} & \textbf{87.38} \\
& \textbf{PERA} ($r=32$) & 0.7012 & \underline{75.54} & \underline{89.83} & 81.58 & \underline{83.53} & \underline{93.56} & \underline{88.40} & 95.61 & 88.95 & \underline{87.12} \\

\midrule
\multirow{3}{*}{Qwen2.5-7B} 
& LoRA ($r=16$) & 0.3541 & 60.00 & 73.60 & 70.00 & 71.70 & 85.90 & 74.40 & 78.60 & 75.80 & 73.80 \\
& HiRA ($r=16$) & 0.3541 & 69.00 & 88.30 & 80.80 & 88.70 & 95.40 & 88.00 & 92.30  & 81.00 & 85.40 \\
& \textbf{PERA} ($r=16$) & 0.3543 & \textbf{73.06} & \textbf{89.61} & \textbf{82.04}&\textbf{88.91} & \textbf{96.17} & \textbf{91.80} & \textbf{95.38} & \textbf{89.34}  & \textbf{88.29} \\

\bottomrule
\end{tabular}}
\caption{Accuracy(\%) comparison of LLaMA model with various fine-tuned methods on commonsense reasoning datasets. Results of all baseline methods are taken from \citep{wu2024mixture, huang2025hira}. The best performance within each LLM is indicated in bold, while the second best performance is highlighted in underline.}
\label{tab:tab1}
\end{table*}

\subsubsection{Relation between LoRA and PERA.}
\label{3.3.3}
We further observe that LoRA can be regarded as a special case of PERA. Concretely, PERA follows the same initialization strategy as LoRA, where the two low-rank matrices $A$ and $B$ are initialized with a Gaussian distribution and zero values. Based on Eq.~\ref{eq8} and Eq.~\ref{eq9}, when the Hadamard vector $\mathbf{h}$ is initialized to zero values and kept frozen during training, denoted as $\{frozen(h_{ij}=0)\mid 1 \le i \le j \le r\}$, PERA becomes equivalent to LoRA. 

Furthermore, when only the cross coefficients are initialized to zero values and kept frozen, represented as $\{frozen(h_{ij})=0)\mid 1 \le i<j\le r  \}$, PERA becomes a LoRA variant that introduces only high-order square feature terms:
\begin{equation}
\label{eq10}
\begin{aligned}
\Delta W 
& = \text{Poly}^{2}(B)\text{Poly}^{2}_{\text{H}}(A) \\
&= \sum_{i=1}^{r} \mathbf{b}_{i} \mathbf{a}_{i}^{T} \\
&\quad + \sum_{1\le i=j}^{r} h_{ij}(\mathbf{b}_{i}\odot \mathbf{b}_{j}) (\mathbf{a}_{i}^{T}\odot \mathbf{a}_{j}^{T}).
\end{aligned}
\end{equation}
Similarly, when only the square coefficients are initialized to zero values and kept frozen, denoted as $\{frozen(h_{ij}=0)\mid 1 \le i =j \le r  \}$, PERA becomes a LoRA variant that incorporates high-order cross feature terms: 
\begin{equation}
\label{eq11}
\begin{aligned}
\Delta W 
& = \text{Poly}^{2}(B)\text{Poly}^{2}_{\text{H}}(A) \\
&= \sum_{i=1}^{r} \mathbf{b}_{i} \mathbf{a}_{i}^{T} \\
&\quad + \sum_{1 \le i < j}^r h_{ij}(\mathbf{b}_{i} \odot \mathbf{b}_{j})(\mathbf{a}_{i}^T \odot \mathbf{a}_{j}^T).
\end{aligned}
\end{equation}
A detailed analysis of the impact of LoRA and its high-order variant can be found in Section~\ref{5.4}

\section{Experiment}
\label{experiment}
In this section, we conduct a systematic evaluation of the PERA across multiple tasks, including commonsense reasoning, natural language understanding(NLU). 
ALL experiments are conducted on NVIDIA RTX 5090 GPUs.
\subsection{Commonsense Reasoning}
\paragraph{Models and Datasets.} We fine-tune LLaMA2-7B~\citep{touvron2023llama2} and LLaMA3-8B~\citep{grattafiori2024llama3} models on the commonsense170K dataset~\citep{hu2023llm-adapters}. This dataset integrates eight commonsense reasoning benchmarks, including BoolQ \citep{clark2019boolq}, PIQA \citep{bisk2020piqa}, SIQA \citep{sap2019socialiqasiqa}, HellaSwag \citep{zellers2019hellaswag}, WinoGrande \citep{sakaguchi2021winogrande}, ARC-c and ARC-e \citep{clark2018thinkarc-e}, and OBQA \citep{mihaylov2018canOBQA}.


\paragraph{Implementation Details.} To ensure a fair comparison, we strictly follow the experimental setup described in~\citep{huang2025hira}. Specifically, when fine-tuning the LLaMA series models, we adjust only the learning rate to $1\text{e}-4$, while keeping all other hyperparameters identical in~\citep{huang2025hira}. For final evaluation, we select the checkpoint that achieves the best performance on the validation set. Detailed training hyperparameters can be found in the Appendix~\ref{Appendix: B.2}.

\paragraph{Main Results.} As shown in Table~\ref{tab:tab1}, PERA consistently demonstrates stable and superior performance across all eight commonsense reasoning tasks on both LLaMA2-7B and LLaMA3-8B. On LLaMA2-7B, PERA achieves an average accuracy of \( 82.61\%\), outperforming LoRA \((77.61\%)\) by \(5\%\). On LLaMA3-8B, PERA reaches \(87.38\%\), surpassing the best baseline method HiRA. These results suggest that PERA provides a more expressive and effective adaptation mechanism for complex reasoning tasks.

Notably, PERA maintains its performance advantages even under extremely low-rank settings. For instance, with rank $r=4$, PERA achieves average accuracies of \(81.57\%\) and \(87.01\%\), which are close to its best performance at $r=16$. This robustness under strict parameter constraints indicates that PERA can effectively exploit low-rank parameters to construct richer adaptation representations, leading to superior performance compared to conventional low-rank adaptation methods.

\begin{table*}[htbp]
\centering
\small
\begin{tabular}{lllccccccc}
\toprule
\textbf{Model} & \textbf{Method} & \textbf{\#Param} & \textbf{SST-2} & \textbf{MRPC} & \textbf{CoLA} & \textbf{QNLI}  & \textbf{RTE} & \textbf{STS-B} & \textbf{Avg.} \\
\midrule
\multirow{6}{*}{RoBERTa-base}
&FFT  & 125M  & 94.40 & 87.90 & 62.40 & 92.50 & 78.30 & 90.60 &  84.40\\
\cmidrule{2-10}
&RED  & 0.02M  & 93.90 & 89.20 & 61.00 & 90.70  & \textbf{78.00} & 90.40 &  83.90\\
&LoRA & 0.3M  & 93.90 & 88.70 & 59.70 & 92.60  & 75.30 & 90.30 &  83.40\\
&HiRA & 0.3M & \textbf{94.22} & \underline{89.53} & 60.30 & 92.39 & 74.68 & 89.58 & 83.45\\
&DeLoRA  & 0.3M  & 94.10 & 89.00 & \underline{63.60} & \underline{92.80} & 77.10 & \underline{90.90} &  \underline{84.60}\\
&\textbf{PERA} & 0.3M & \underline{94.13} & \textbf{89.71} & \textbf{64.80} & \textbf{92.94} & \underline{77.84} & \textbf{91.13} & \textbf{85.10}\\
\midrule
\multirow{6}{*}{RoBERTa-large}
&FFT  & 355M  & 96.00 & 91.70 & 68.20 & 93.80 & 85.80 & 92.60 & 88.00 \\
\cmidrule{2-10}
&RED  & 0.05M  & \underline{96.00} & \underline{90.30} & \underline{68.10} & 93.50  & 86.20 & 91.30 & \underline{87.60} \\
&LoRA & 0.8M  & \underline{96.00} & 89.80 & 65.50 & \underline{94.70}  & \underline{86.30} & \underline{91.70} & 87.30 \\
&\textbf{PERA} & 0.8M & \textbf{96.24} & \textbf{90.78} & \textbf{68.40} &\textbf{94.89} & \textbf{86.47} & \textbf{91.98} & \textbf{88.13}\\
\bottomrule
\end{tabular}
\caption{Comparisons of different methods finetuning RoBERTa-base and RoBERTa-large on GLUE benchmark. Results of all baselines are taken from~\citep{bini2025delora, wu2024reft} except HiRA. The best performance within each LLM is indicated in bold, while the second best performance is highlighted in underline.}
\label{tab:tab2}
\end{table*}

\subsection{Natural Language Understanding}
\paragraph{Models and Datasets.} We evaluate the effectiveness of PERA for adapting small language models by fine-tuning the RoBERTa-base model~\citep{roberta_base_large} on the GLUE (General Language Understanding Evaluation) benchmark~\citep{wang2018glue}. The GLUE benchmark is widely used to assess natural language understanding capabilities and comprises a diverse set of tasks, including natural language inference, sentiment classification and textual entailment. We conduct experiment on six datasets from the GLUE benchmark: SST-2, MRPC, CoLA, QNLI, RTE, and STS-B. Following the experimental setup of~\citep{wu2024advancing,wu2024reft,bini2025delora}, we randomly split the validation set into two subsets using a pre-defined seed: one subset for model hyperparameter selection and the other is reserved for final performance evaluation. Detailed statistics for each dataset are provided in Appendix~\ref{Appendix: B.1}
\paragraph{Implementation Details.} We follow the experimental setup used in~\citep{wu2024advancing,wu2024reft,bini2025delora} to ensure fair comparison across various methods. To maintain a consistent number of trainable parameters among different approaches, we fine-tune only the query (Q) and value (V) projection layers in the attention modules and fix the rank to $r=8$ for all datasets. ALL reported results are averaged over five runs with different random seeds. Additional implementation and hyperparameter details can be found in Appendix~\ref{Appendix: B.2}
\paragraph{Main Results.} 
Table~\ref{tab:tab2} reports the NLU results on six benchmark datasets using different RoBERTa models. PERA consistently outperforms all existing PEFT methods across every evaluated configuration.
Notably, under the same rank setting ($r=8$), PERA achieves average accuracy gains of $1.70\%$ and $0.83\%$ over LoRA on RoBERTa-base and RoBERTa-large, respectively.
Moreover, on the RoBERTa-large model, PERA delivers the best performance on all six datasets, demonstrating its strong generalization and the effectiveness of high-order nonlinear feature modeling even in high-capacity models.


\section{Understanding the PERA}

\subsection{Impact on the Number of Rank}
\label{5.1}
We conduct a systematic study to investigate how the parameter rank $r$ influences model performance across multiple commonsense reasoning tasks. In the experiments, we vary only the value of $r$ while keeping all other hyperparameters fixed and evaluate the LLaMA3-8B model using the average accuracy as the evaluation metric. The results, as shown in Figure~\ref{fig:fig3}, reveal several important observations:
\begin{figure}
    \centering
    \includegraphics[width=1.0\linewidth]{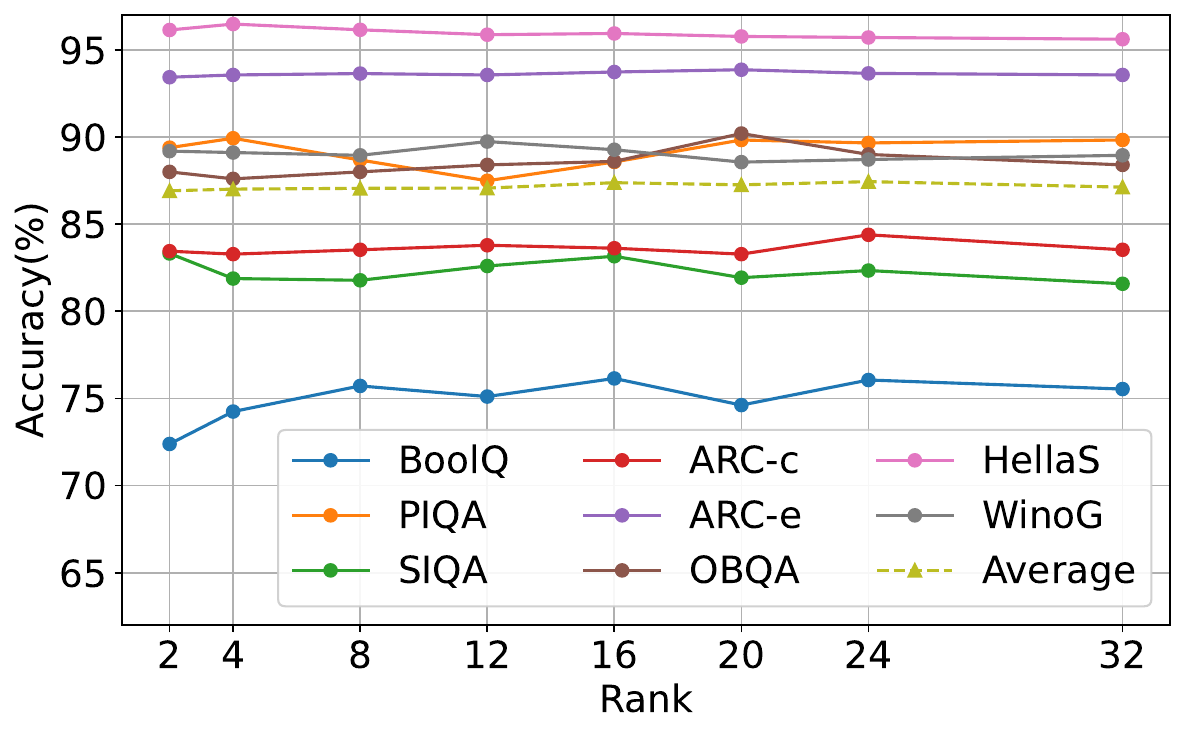}
    \caption{Accuracy\((\%)\) comparison with the rank $r$ increases on LLaMA3-8B model. The detailed results are provided in Appendix~\ref{Appendix: C.1}}
    \label{fig:fig3}
\end{figure}
First, even under extremely low-rank settings ($r=2$ and $r=4$), PERA achieves strong performance, with average accuracy of $86.91\%$ and $87.01\%$. This indicates that PERA can effectively leverage low-rank parameters to construct high-order interaction terms, thereby enriching adaptation representations. Second, as the rank increases from $2$ to $16$, performance improves substantially, with the average accuracy rising from \(86.91\%\) to \(87.38\%\), which corresponds to the best observed result.  These observations suggest that PERA can achieve performance comparable to high-resource configurations even in low-resource settings. This rank robustness can be attributed primarily to the polynomial expansion mechanism, which effectively increases the maximum achievable rank of updated weights and enables high-order interactions within the low-rank space, thereby enhancing the expressive capacity of the model. 

\begin{table}[h]
\centering
\setlength{\tabcolsep}{13pt}
\resizebox{\linewidth}{!}{
\begin{tabular}{l c c}
\toprule
\textbf{Placement} &  \textbf{\#Params(\%)} & \textbf{Avg.} \\
\midrule
QKV, UD & 0.3513 &\textbf{87.38} \\
QKV & 0.1176 & 87.02 \\
UD & 0.2345& 86.91 \\
QK & 0.0849& 85.25 \\
QV & 0.0849& 86.70 \\
Q  & 0.0522& 84.83 \\
K  & 0.0326& 83.53 \\
V  & 0.0326& 86.19 \\
\bottomrule
\end{tabular}}
\caption{Average accuracy (\%) comparison of different modules on various tasks with LLaMA3-8B model.}
\label{tab:tab3}
\end{table}
\subsection{Impact on Placement in Transformers}
\label{5.2}
In this section, we analyze the effect of applying PERA to different weight modules within the Transformer architecture. Each module is denoted by its first letter: (Q)uery, (K)ey, (V)alue, (U)p, and (D)own. As shown in Table~\ref{tab:tab3}, the experimental results show that applying PERA jointly to both the QKV and UD layers yields the best performance, which is consistent with the structure recommended by LoRA. In contrast, restricting PERA to a single module leads to a degradation in overall performance. Furthermore, adapting the QKV layers consistently outperforms adapting only the UD layers. The detailed results can be found in Appendix~\ref{Appendix: C.2}

\begin{figure}
    \centering
    \includegraphics[width=0.96\linewidth]{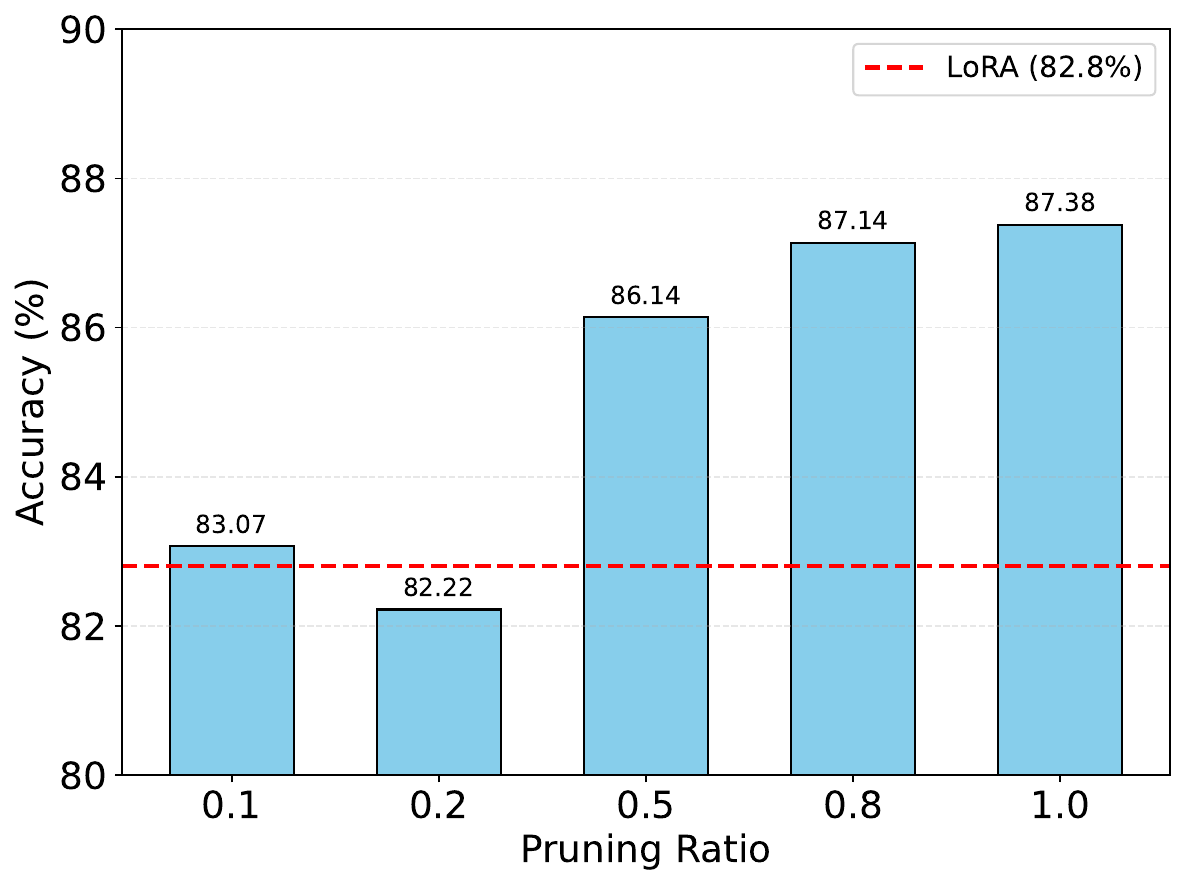}
    \caption{Average accuracy under different pruning ratios of Commonsense170K with LLaMA3-8B model.}
    \label{fig:fig4}
\end{figure}

\begin{table}[h]
\centering
\setlength{\tabcolsep}{13pt}
\resizebox{\linewidth}{!}{
\begin{tabular}{l c c }
\toprule
\textbf{Method} & \(\Delta W\) & \textbf{Avg.} \\
\midrule
LoRA  & Eq.~\ref{eq8} &  82.80\\
LoRA + square-only & Eq.~\ref{eq10} & 87.48 \\
LoRA + cross-only & Eq.~\ref{eq11} &  86.83\\
PERA  & Eq.~\ref{eq9} & 87.38 \\
\bottomrule
\end{tabular}}
\caption{Ablation of PERA innovations on the commonsense reasoning tasks. We show how different components affect performance from LoRA~\citep{hu2022lora} with the LLaMA3-8B model.}
\label{tab:tab4}
\end{table}
\subsection{Impact on Low Resources}
\label{section 5.3}
To evaluate the performance of PERA under low-resource conditions, we partition the commonsense170k dataset and randomly sample \(10\%\), \(20\%\), \(50\%\) and \(80\%\) of the data for experimentation. Figure~\ref{fig:fig4} illustrates the relationship between training dataset pruning ratio and average accuracy, with detailed numerical results reported in the Appendix~\ref{Appendix: C.3}. The results demonstrate that even when trained on only \(10\%\) of the dataset, PERA achieves an average accuracy of \(83.07\%\), surpassing LoRA \((82.80\%)\) trained on the full commonsense170K dataset. Moreover, as the number of training datasets increases, the performance gap between PERA and LoRA widens, highlighting its data efficiency in low-resource settings.

\subsection{Impact on Different High-Order Components with LoRA}
\label{5.4}
Since LoRA can be viewed as a special case of PERA under specific initialization, we further examine the individual effects of introducing high-order square and cross feature terms. The standard LoRA formulation is shown in Eq.~\ref{eq8}, while the square-only and cross-only variants are defined in Eq.~\ref{eq10} and Eq.~\ref{eq11}, respectively. PERA (Eq.~\ref{eq9}) can thus be regarded as a generalized LoRA variant that jointly incorporates both types of high-order terms. The result is shown in Table~\ref{tab:tab4}.

Experiments on commonsense reasoning tasks using LLaMA3-8B (rank = 16) show that incorporating either square ($87.48\%$) or cross ($86.83\%$) terms notably improves performance over LoRA ($82.80\%$). Among them, the square feature expansion yields the largest gain, highlighting its importance in enhancing model expressivity. Combining both high-order terms (PERA) yields a comparable improvement ($87.38\%$), suggesting that excessive interaction modeling may introduce redundancy.

\begin{figure}[h]
    \centering
    \begin{subfigure}{0.85\linewidth}
        \centering
        \includegraphics[width=\linewidth]{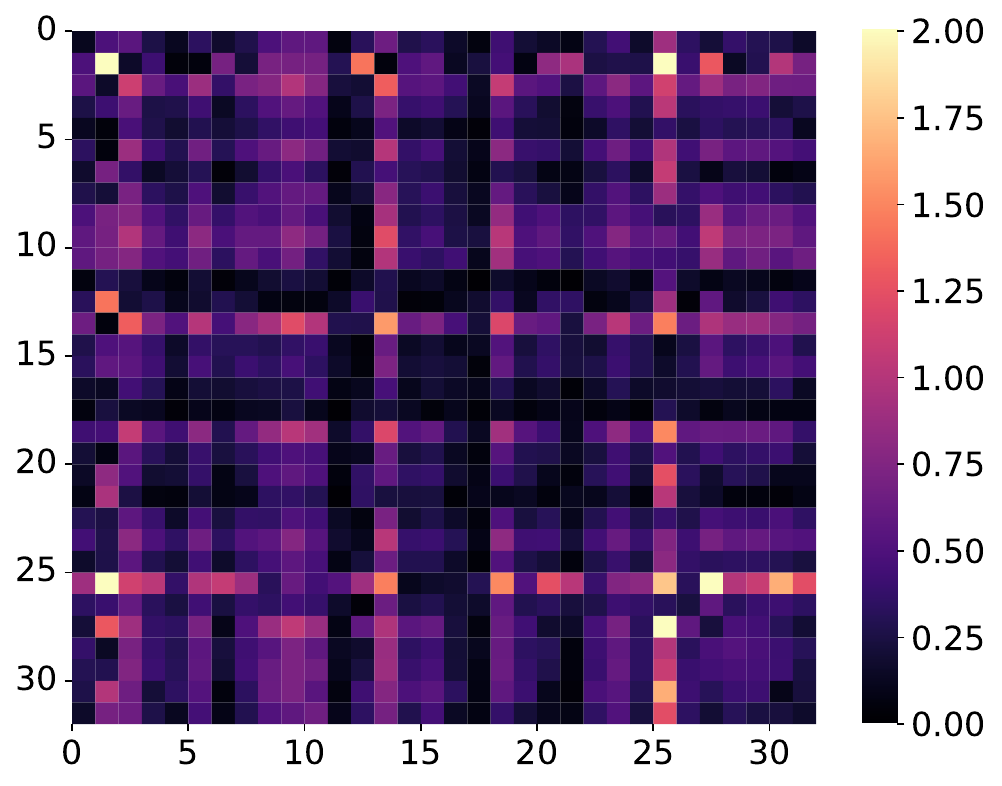}
        \caption{LoRA.}
    \end{subfigure}
    \begin{subfigure}{0.85\linewidth}
        \centering
        \includegraphics[width=\linewidth]{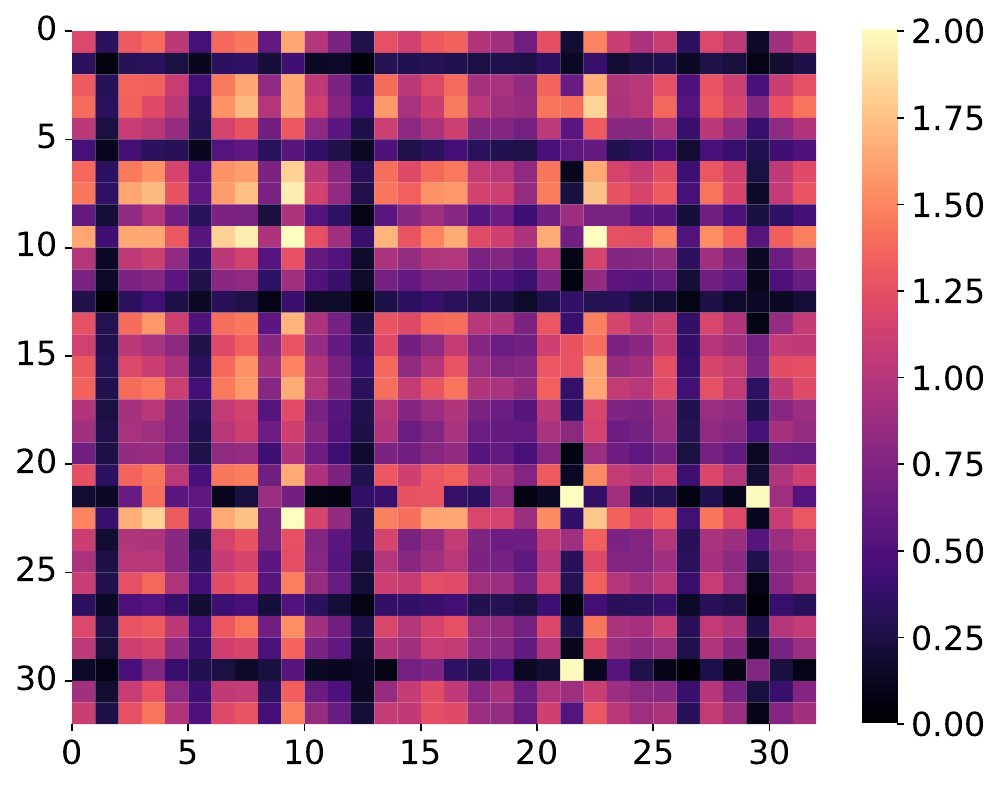}
        \caption{PERA.}
    \end{subfigure}
    \caption{The interaction strength matrix of LoRA and PERA. We quantify the combined influence of each feature dimension on the output, not just the individual influence.}
    \label{fig:fig5}
\end{figure}


\subsection{Feature Interaction Strength Analysis}
We compare our method with LoRA using the Hessian-based Interaction Strength Matrix. Specifically, for the model output $f(h)$, we compute the absolute second-order partial derivatives with respect to the hidden representation: 
\begin{equation}
S_{ij} = \mathbb{E}_{h} \left[ \left| \frac{\partial^2 f(h)}{\partial h_i \partial h_j} \right| \right].
\end{equation}
As shown in Figure~\ref{fig:fig5}, we randomly select the slices of a small set of hidden states. Our method yields a higher global interaction strength compared to LoRA, indicating a more expressive modeling of high-order feature coupling.
\subsection{Memory and Speed Analysis}
\begin{table}[t]
\centering
\resizebox{\linewidth}{!}{
\begin{tabular}{l cc cc}
\toprule
\multirow{2}{*}{Method}
& \multicolumn{2}{c}{Training} 
& \multicolumn{2}{c}{Inference} \\
\cmidrule(lr){2-3} \cmidrule(lr){4-5}
 & Memory & Speed & Memory & Speed \\
\midrule
LoRA  & 18.70GB & 12h 27m & 19.50GB & 9m30s \\
DoRA  & 19.58GB & 22h 07m & 19.94GB & 23m40s \\
HiRA  & 18.72GB & 14h 02m & 19.58GB & 20m45s \\
PERA  & 19.12GB & 13h 30m & 19.70GB & 14m53s \\
\bottomrule
\end{tabular}}
\caption{Comparison of training and inference efficiency across methods.}
\label{tab:tab5}
\end{table}

To provide a more comparison on training and inference cost, we compare PERA with LoRA, DoRA, and HiRA under identical settings (Commonsense170K, rank=4, batch size=16, LLaMA2-7B, 3 epochs and evaluating on BoolQ dataset) on the same single RTX5090 32G GPU. We report peak memory usage and total training time. The results are summarized in Table~\ref{tab:tab5}.

PERA implements higher-order interaction terms via matrix concatenation rather than sequential matrix additions, which avoids additional forward passes and limits computational overhead. Overall, the results indicate that PERA maintains a computational and memory footprint close to standard LoRA and remain substantially more efficient than DoRA.

\section{Conclusion}
We introduce PERA, a polynomial expansion–based low-rank adaptation method that enables higher-order parameter interactions while preserving efficiency. Our analysis and experiments highlight the importance of high-order nonlinear components, especially square terms, in improving the expressive capacity of low-rank fine-tuning. These results indicate that structured modeling of higher-order parameter relationships offers a promising direction for more expressive and efficient adaptation.
\section{Limitations}
In this work, we primarily evaluate PERA on commonsense reasoning tasks and GLUE benchmarks using LLaMA and RoBERTa models. While these results demonstrate the effectiveness of PERA in language understanding and reasoning scenarios, the current evaluation does not cover other domains that may require different forms of adaptation. In particular, tasks such as arithmetic reasoning or multimodal generation (e.g., image generation) may exhibit distinct characteristics and challenges. Exploring the applicability and effectiveness of PERA in these domains remains an important direction for future work.

\section{Ethics Statement}
This work introduces PERA, a parameter-efficient fine-tuning method that operates entirely in the parameter space of pretrained language models. The research does not involve the collection or annotation of new data, human subjects, or user-generated content. All datasets used in our experiments are publicly available and widely adopted in prior PEFT studies.
The proposed method aims to improve model expressivity and efficiency without altering model behavior or generating new textual content. Therefore, it poses no additional ethical or societal risks beyond standard fine-tuning techniques. We encourage responsible use of PEFT methods, particularly regarding dataset bias and downstream deployment contexts.

\section{Acknowledgements}
This work is supported by the National Natural Science Foundation of China (No.62206004, No.62572002, No.62272001, No.624065095), and the Natural Science Foundation of Anhui Province (No.2208085QF199, No.2508085MF159, No.2308085MF213).

\bibliography{custom}

@article{brown2020language,
  title={Language models are few-shot learners},
  author={Brown, Tom and Mann, Benjamin and Ryder, Nick and Subbiah, Melanie and Kaplan, Jared D and Dhariwal, Prafulla and Neelakantan, Arvind and Shyam, Pranav and Sastry, Girish and Askell, Amanda and others},
  journal={Advances in neural information processing systems},
  volume={33},
  pages={1877--1901},
  year={2020}
}

@article{touvron2023llama,
  title={Llama 2: Open foundation and fine-tuned chat models},
  author={Touvron, Hugo and Martin, Louis and Stone, Kevin and Albert, Peter and Almahairi, Amjad and Babaei, Yasmine and Bashlykov, Nikolay and Batra, Soumya and Bhargava, Prajjwal and Bhosale, Shruti and others},
  journal={arXiv preprint arXiv:2307.09288},
  year={2023}
}

@article{touvron2023llama2,
  title={Llama: Open and efficient foundation language models},
  author={Touvron, Hugo and Lavril, Thibaut and Izacard, Gautier and Martinet, Xavier and Lachaux, Marie-Anne and Lacroix, Timoth{\'e}e and Rozi{\`e}re, Baptiste and Goyal, Naman and Hambro, Eric and Azhar, Faisal and others},
  journal={arXiv preprint arXiv:2302.13971},
  year={2023}
}

@article{ouyang2022training,
  title={Training language models to follow instructions with human feedback},
  author={Ouyang, Long and Wu, Jeffrey and Jiang, Xu and Almeida, Diogo and Wainwright, Carroll and Mishkin, Pamela and Zhang, Chong and Agarwal, Sandhini and Slama, Katarina and Ray, Alex and others},
  journal={Advances in neural information processing systems},
  volume={35},
  pages={27730--27744},
  year={2022}
}

@inproceedings{ruiz2023dreambooth,
  title={Dreambooth: Fine tuning text-to-image diffusion models for subject-driven generation},
  author={Ruiz, Nataniel and Li, Yuanzhen and Jampani, Varun and Pritch, Yael and Rubinstein, Michael and Aberman, Kfir},
  booktitle={Proceedings of the IEEE/CVF conference on computer vision and pattern recognition},
  pages={22500--22510},
  year={2023}
}

@article{thirunavukarasu2023large,
  title={Large language models in medicine},
  author={Thirunavukarasu, Arun James and Ting, Darren Shu Jeng and Elangovan, Kabilan and Gutierrez, Laura and Tan, Ting Fang and Ting, Daniel Shu Wei},
  journal={Nature medicine},
  volume={29},
  number={8},
  pages={1930--1940},
  year={2023},
  publisher={Nature Publishing Group US New York}
}

@inproceedings{houlsby2019parameter,
  title={Parameter-efficient transfer learning for NLP},
  author={Houlsby, Neil and Giurgiu, Andrei and Jastrzebski, Stanislaw and Morrone, Bruna and De Laroussilhe, Quentin and Gesmundo, Andrea and Attariyan, Mona and Gelly, Sylvain},
  booktitle={International conference on machine learning},
  pages={2790--2799},
  year={2019},
  organization={PMLR}
}

@article{lialin2023scaling,
  title={Scaling down to scale up: A guide to parameter-efficient fine-tuning},
  author={Lialin, Vladislav and Deshpande, Vijeta and Rumshisky, Anna},
  journal={arXiv preprint arXiv:2303.15647},
  year={2023}
}

@article{hu2022lora,
  title={Lora: Low-rank adaptation of large language models.},
  author={Hu, Edward J and Shen, Yelong and Wallis, Phillip and Allen-Zhu, Zeyuan and Li, Yuanzhi and Wang, Shean and Wang, Lu and Chen, Weizhu and others},
  journal={ICLR},
  volume={1},
  number={2},
  pages={3},
  year={2022}
}

@article{liu2021p,
  title={P-tuning v2: Prompt tuning can be comparable to fine-tuning universally across scales and tasks},
  author={Liu, Xiao and Ji, Kaixuan and Fu, Yicheng and Tam, Weng Lam and Du, Zhengxiao and Yang, Zhilin and Tang, Jie},
  journal={arXiv preprint arXiv:2110.07602},
  year={2021}
}

@article{jiang2024mora,
  title={Mora: High-rank updating for parameter-efficient fine-tuning},
  author={Jiang, Ting and Huang, Shaohan and Luo, Shengyue and Zhang, Zihan and Huang, Haizhen and Wei, Furu and Deng, Weiwei and Sun, Feng and Zhang, Qi and Wang, Deqing and others},
  journal={arXiv preprint arXiv:2405.12130},
  year={2024}
}

@article{zhuang2024time,
  title={Time-Varying LoRA: Towards effective cross-domain fine-tuning of diffusion models},
  author={Zhuang, Zhan and Zhang, Yulong and Wang, Xuehao and Lu, Jiangang and Wei, Ying and Zhang, Yu},
  journal={Advances in Neural Information Processing Systems},
  volume={37},
  pages={73920--73951},
  year={2024}
}

@article{liu2023chipnemo,
  title={Chipnemo: Domain-adapted llms for chip design},
  author={Liu, Mingjie and Ene, Teodor-Dumitru and Kirby, Robert and Cheng, Chris and Pinckney, Nathaniel and Liang, Rongjian and Alben, Jonah and Anand, Himyanshu and Banerjee, Sanmitra and Bayraktaroglu, Ismet and others},
  journal={arXiv preprint arXiv:2311.00176},
  year={2023}
}

@article{hu2023llm-adapters,
  title={Llm-adapters: An adapter family for parameter-efficient fine-tuning of large language models},
  author={Hu, Zhiqiang and Wang, Lei and Lan, Yihuai and Xu, Wanyu and Lim, Ee-Peng and Bing, Lidong and Xu, Xing and Poria, Soujanya and Lee, Roy Ka-Wei},
  journal={arXiv preprint arXiv:2304.01933},
  year={2023}
}

@article{clark2019boolq,
  title={Boolq: Exploring the surprising difficulty of natural yes/no questions},
  author={Clark, Christopher and Lee, Kenton and Chang, Ming-Wei and Kwiatkowski, Tom and Collins, Michael and Toutanova, Kristina},
  journal={arXiv preprint arXiv:1905.10044},
  year={2019}
}

@inproceedings{bisk2020piqa,
  title={Piqa: Reasoning about physical commonsense in natural language},
  author={Bisk, Yonatan and Zellers, Rowan and Gao, Jianfeng and Choi, Yejin and others},
  booktitle={Proceedings of the AAAI conference on artificial intelligence},
  volume={34},
  number={05},
  pages={7432--7439},
  year={2020}
}

@article{sap2019socialiqasiqa,
  title={Socialiqa: Commonsense reasoning about social interactions},
  author={Sap, Maarten and Rashkin, Hannah and Chen, Derek and LeBras, Ronan and Choi, Yejin},
  journal={arXiv preprint arXiv:1904.09728},
  year={2019}
}

@article{zellers2019hellaswag,
  title={Hellaswag: Can a machine really finish your sentence?},
  author={Zellers, Rowan and Holtzman, Ari and Bisk, Yonatan and Farhadi, Ali and Choi, Yejin},
  journal={arXiv preprint arXiv:1905.07830},
  year={2019}
}

@article{sakaguchi2021winogrande,
  title={Winogrande: An adversarial winograd schema challenge at scale},
  author={Sakaguchi, Keisuke and Bras, Ronan Le and Bhagavatula, Chandra and Choi, Yejin},
  journal={Communications of the ACM},
  volume={64},
  number={9},
  pages={99--106},
  year={2021},
  publisher={ACM New York, NY, USA}
}

@article{clark2018thinkarc-e,
  title={Think you have solved question answering? try arc, the ai2 reasoning challenge},
  author={Clark, Peter and Cowhey, Isaac and Etzioni, Oren and Khot, Tushar and Sabharwal, Ashish and Schoenick, Carissa and Tafjord, Oyvind},
  journal={arXiv preprint arXiv:1803.05457},
  year={2018}
}

@article{mihaylov2018canOBQA,
  title={Can a suit of armor conduct electricity? a new dataset for open book question answering},
  author={Mihaylov, Todor and Clark, Peter and Khot, Tushar and Sabharwal, Ashish},
  journal={arXiv preprint arXiv:1809.02789},
  year={2018}
}

@article{grattafiori2024llama3,
  title={The llama 3 herd of models},
  author={Grattafiori, Aaron and Dubey, Abhimanyu and Jauhri, Abhinav and Pandey, Abhinav and Kadian, Abhishek and Al-Dahle, Ahmad and Letman, Aiesha and Mathur, Akhil and Schelten, Alan and Vaughan, Alex and others},
  journal={arXiv preprint arXiv:2407.21783},
  year={2024}
}

@inproceedings{huang2025hira,
  title={HiRA: Parameter-efficient hadamard high-rank adaptation for large language models},
  author={Huang, Qiushi and Ko, Tom and Zhuang, Zhan and Tang, Lilian and Zhang, Yu},
  booktitle={The Thirteenth International Conference on Learning Representations},
  year={2025}
}

@article{li2021prefix,
  title={Prefix-tuning: Optimizing continuous prompts for generation},
  author={Li, Xiang Lisa and Liang, Percy},
  journal={arXiv preprint arXiv:2101.00190},
  year={2021}
}

@article{lester2021power,
  title={The power of scale for parameter-efficient prompt tuning},
  author={Lester, Brian and Al-Rfou, Rami and Constant, Noah},
  journal={arXiv preprint arXiv:2104.08691},
  year={2021}
}

@inproceedings{liu2024dora,
  title={Dora: Weight-decomposed low-rank adaptation},
  author={Liu, Shih-Yang and Wang, Chien-Yi and Yin, Hongxu and Molchanov, Pavlo and Wang, Yu-Chiang Frank and Cheng, Kwang-Ting and Chen, Min-Hung},
  booktitle={Forty-first International Conference on Machine Learning},
  year={2024}
}

@article{bini2025delora,
  title={DeLoRA: Decoupling Angles and Strength in Low-rank Adaptation},
  author={Bini, Massimo and Girrbach, Leander and Akata, Zeynep},
  journal={arXiv preprint arXiv:2503.18225},
  year={2025}
}

@misc{mistral7b,
      title={Mistral 7B}, 
      author={Albert Q. Jiang and Alexandre Sablayrolles and Arthur Mensch and Chris Bamford and Devendra Singh Chaplot and Diego de las Casas and Florian Bressand and Gianna Lengyel and Guillaume Lample and Lucile Saulnier and Lélio Renard Lavaud and Marie-Anne Lachaux and Pierre Stock and Teven Le Scao and Thibaut Lavril and Thomas Wang and Timothée Lacroix and William El Sayed},
      year={2023},
      eprint={2310.06825},
      archivePrefix={arXiv},
      primaryClass={cs.CL},
      url={https://arxiv.org/abs/2310.06825}, 
}

@inproceedings{wu2024advancing,
  title={Advancing parameter efficiency in fine-tuning via representation editing},
  author={Wu, Muling and Liu, Wenhao and Wang, Xiaohua and Li, Tianlong and Lv, Changze and Ling, Zixuan and JianHao, Zhu and Zhang, Cenyuan and Zheng, Xiaoqing and Huang, Xuan-Jing},
  booktitle={Proceedings of the 62nd Annual Meeting of the Association for Computational Linguistics (Volume 1: Long Papers)},
  pages={13445--13464},
  year={2024}
}

@article{wu2024reft,
  title={Reft: Representation finetuning for language models},
  author={Wu, Zhengxuan and Arora, Aryaman and Wang, Zheng and Geiger, Atticus and Jurafsky, Dan and Manning, Christopher D and Potts, Christopher},
  journal={Advances in Neural Information Processing Systems},
  volume={37},
  pages={63908--63962},
  year={2024}
}

@inproceedings{roberta_base_large,
  title={A robustly optimized BERT pre-training approach with post-training},
  author={Zhuang, Liu and Wayne, Lin and Ya, Shi and Jun, Zhao},
  booktitle={Proceedings of the 20th chinese national conference on computational linguistics},
  pages={1218--1227},
  year={2021}
}

@inproceedings{wang2018glue,
  title={GLUE: A multi-task benchmark and analysis platform for natural language understanding},
  author={Wang, Alex and Singh, Amanpreet and Michael, Julian and Hill, Felix and Levy, Omer and Bowman, Samuel},
  booktitle={Proceedings of the 2018 EMNLP workshop BlackboxNLP: Analyzing and interpreting neural networks for NLP},
  pages={353--355},
  year={2018}
}

@article{polynomial_expansion1,
  title={Python Feature Engineering Cookbook: Over 70 Recipes for Creating},
  author={Galli, Soledad},
  journal={Engineering, and Transforming Features to Build Machine Learning Models},
  year={2020}
}

@book{polynomial_expansion2,
  title={Feature engineering and selection: A practical approach for predictive models},
  author={Kuhn, Max and Johnson, Kjell},
  year={2019},
  publisher={Chapman and Hall/CRC}
}

@book{polynomial_expansion3,
  title={Feature engineering bookcamp},
  author={Ozdemir, Sinan},
  year={2022},
  publisher={Simon and Schuster}
}

@inproceedings{williams2018broad,
  title={A broad-coverage challenge corpus for sentence understanding through inference},
  author={Williams, Adina and Nangia, Nikita and Bowman, Samuel},
  booktitle={Proceedings of the 2018 conference of the North American chapter of the association for computational linguistics: human language technologies, volume 1 (long papers)},
  pages={1112--1122},
  year={2018}
}

@inproceedings{socher2013recursive,
  title={Recursive deep models for semantic compositionality over a sentiment treebank},
  author={Socher, Richard and Perelygin, Alex and Wu, Jean and Chuang, Jason and Manning, Christopher D and Ng, Andrew Y and Potts, Christopher},
  booktitle={Proceedings of the 2013 conference on empirical methods in natural language processing},
  pages={1631--1642},
  year={2013}
}

@inproceedings{dolan2005automatically,
  title={Automatically constructing a corpus of sentential paraphrases},
  author={Dolan, William B and Brockett, Chris},
  booktitle={Proceedings of the third international workshop on paraphrasing (IWP2005)},
  year={2005}
}

@article{cer2017semeval,
  title={Semeval-2017 task 1: Semantic textual similarity-multilingual and cross-lingual focused evaluation},
  author={Cer, Daniel and Diab, Mona and Agirre, Eneko and Lopez-Gazpio, Inigo and Specia, Lucia},
  journal={arXiv preprint arXiv:1708.00055},
  year={2017}
}

@article{demszky2018transforming,
  title={Transforming question answering datasets into natural language inference datasets},
  author={Demszky, Dorottya and Guu, Kelvin and Liang, Percy},
  journal={arXiv preprint arXiv:1809.02922},
  year={2018}
}

@inproceedings{haim2006second,
  title={The second pascal recognising textual entailment challenge},
  author={Haim, R Bar and Dagan, Ido and Dolan, Bill and Ferro, Lisa and Giampiccolo, Danilo and Magnini, Bernardo and Szpektor, Idan},
  booktitle={Proceedings of the Second PASCAL Challenges Workshop on Recognising Textual Entailment},
  volume={7},
  pages={785--794},
  year={2006}
}

@article{warstadt2019neural,
  title={Neural network acceptability judgments},
  author={Warstadt, Alex and Singh, Amanpreet and Bowman, Samuel R},
  journal={Transactions of the Association for Computational Linguistics},
  volume={7},
  pages={625--641},
  year={2019},
  publisher={MIT Press One Rogers Street, Cambridge, MA 02142-1209, USA journals-info~…}
}

@article{Attention,
  title={Attention is all you need},
  author={Vaswani, Ashish and Shazeer, Noam and Parmar, Niki and Uszkoreit, Jakob and Jones, Llion and Gomez, Aidan N and Kaiser, {\L}ukasz and Polosukhin, Illia},
  journal={Advances in neural information processing systems},
  volume={30},
  year={2017}
}

@article{wu2024mixture,
  title={Mixture-of-subspaces in low-rank adaptation},
  author={Wu, Taiqiang and Wang, Jiahao and Zhao, Zhe and Wong, Ngai},
  journal={arXiv preprint arXiv:2406.11909},
  year={2024}
}

@article{eckart1936approximation,
  title={The approximation of one matrix by another of lower rank},
  author={Eckart, Carl and Young, Gale},
  journal={Psychometrika},
  volume={1},
  number={3},
  pages={211--218},
  year={1936},
  publisher={Springer-Verlag}
}
\appendix

\onecolumn
\section{Analysis Proof}
\subsection{Rank Analysis}
\label{Appendix: A.1}
\paragraph{LoRA.} The weight update matrix $\Delta W$ can be decomposed into the product of two low-rank matrices $A\in \mathbb{R}^{r\times n}$ and $B\in \mathbb{R}^{m\times r}$. Given a pretrained weight matrix $W_{0}\in \mathbb{R}^{m\times n}$ with rank $r_{0}$, the rank of the final weight matrix $W'$ with LoRA satisfies the following relationship:

\begin{equation*}
\begin{aligned}
    \mathrm{Rank}(W') = \mathrm{Rank}(W_{0} + \Delta W) = \mathrm{Rank}(W_{0} + BA).
\end{aligned}
\end{equation*}
According to the properties of rank in a matrix, the upper bound of LoRA can be expressed as follows:
\begin{equation*}
\begin{aligned}
    \mathrm{Rank}(W_{0} + BA) 
    & \leq \mathrm{Rank}(W_{0}) + \mathrm{Rank}(BA) \\
    & \leq \mathrm{Rank}(W_{0}) + \min(\mathrm{Rank}(B), \mathrm{Rank}(A))\\
    & \leq r_{0} + r.
\end{aligned}
\end{equation*}

\paragraph{PERA.} PERA first applies the polynomial expansion to the low-rank matrix $B \in \mathbb{R}^{m\times r}$ to obtain a high-rank matrix $\hat{B}\in \mathbb{R}^{m\times (2r+C(r,2))}$. Meanwhile, the polynomial expansion with Hadamard product is applied to the low-rank matrix $A\in \mathbb{R}^{r\times n}$ to generate a high-rank matrix $\hat{A}\in \mathbb{R}^{(2r+C(r,2))\times n}$. The weight update matrix can be decomposed as the product of these two high-rank matrices. The rank of the final weight matrix $W'$ with PERA satisfies the following relationship: 

\begin{equation*}
\begin{aligned}
    \mathrm{Rank}(W') = \mathrm{Rank}(W_{0} + \Delta W) = \mathrm{Rank}\!\left(W_0+\mathrm{Poly}^{2}(B)\,\mathrm{Poly}^{2}_{H}(A)\right).
\end{aligned}
\end{equation*}
According to the properties of rank in a matrix, the upper bound of PERA can be expressed as follows:
\begin{equation*} 
\begin{aligned}
    \mathrm{Rank}\!\left(W_0+\mathrm{Poly}^{2}(B)\,\mathrm{Poly}^{2}_{H}(A)\right)
    &\leq \mathrm{Rank}(W_{0}) + \mathrm{Rank}\!\left(\mathrm{Poly}^{2}(B)\,\mathrm{Poly}^{2}_{H}(A)\right) \\
    &\leq \mathrm{Rank}(W_{0}) 
        + \min\!\left(\mathrm{Rank}\!\left(\mathrm{Poly}^{2}(B)\right), 
        \mathrm{Rank}\!\left(\mathrm{Poly}^{2}_{H}(A)\right)\right) \\
    &\leq r_{0} + (2r+C(r,2)).
\end{aligned}
\end{equation*}

\subsection{Feature Utilization Analysis}
\label{Appendix: A.2}
\paragraph{LoRA.} In LoRA, the weight update matrix is constructed by the product of two low-rank matrices $B\in \mathbb{R}^{m\times r}$ and $A\in \mathbb{R}^{r\times n}$: 
\begin{equation*}
    \Delta W = BA .
\end{equation*}
The low-rank matrices $B$ and $A$ can be represented in terms of column and row vectors, respectively:
\begin{equation*}
    B = \left[\, \mathbf{b}_i \;\middle|\; 1\leq i \leq r  \,\right];
\end{equation*}
\begin{equation*}
    A = \left[\, \mathbf{a}_i \;\middle|\; 1\leq i \leq r  \,\right]^T ,
\end{equation*}
where each $\mathbf{b}_{i} \in \mathbb{R}^{m}$ corresponds to a column vector in $B$ and each $\mathbf{a}_{i} \in \mathbb{R}^{n}$ corresponds to a row vector in $A$. Therefore, the weight update matrix $\Delta W$ of LoRA can be expressed as:

\begin{equation*}
    \Delta W = BA = \sum_{i=1}^{r}\mathbf{b}_{i}\mathbf{a}_{i}^T.
\end{equation*}
\paragraph{PERA.} PERA decomposes the weight update matrix into the product of two high-rank matrices, $\hat{B}\in \mathbb{R}^{m\times (2r+C(r,2))}$ and $\hat{A}\in \mathbb{R}^{ (2r+C(r,2))\times n}$:

\begin{equation*}
    \Delta W = \hat{B}\hat{A} = \text{Poly}^{2}{(B)\text{Poly}^{2}_{\text{H}}(A)}
\end{equation*}
The low-rank matrices $\hat{B}$ and $\hat{A}$ can be represented in terms of column and row vectors, respectively:
\begin{equation*}
\text{Poly}^{2}(B)=
\left[
\begin{array}{l}
\mathbf{b}_i; \\
\mathbf{b}_i \odot \mathbf{b}_j; \\
\mathbf{b}_i \odot \mathbf{b}_j
\end{array}
\;\middle|\;
\begin{array}{l}
1 \le i \le r;\\
1 \le i = j \le r;\\
1 \le i < j \le r
\end{array}
\right].
\end{equation*}

\begin{equation*}
\text{Poly}^{2}_{\text{H}}(A) =
\left[
\begin{array}{l}
\mathbf{{a}_i}; \\
h_{ij}(\mathbf{a}_i \odot \mathbf{a}_j); \\
h_{ij}(\mathbf{a}_i \odot \mathbf{a}_j)
\end{array}
\;\middle|\;
\begin{array}{l}
1 \le i \le r;\\
1 \le i =j\le r;\\
1 \le i < j \le r
\end{array}\right]^T.
\end{equation*}
Therefore, the weight update matrix $\Delta W$ of PERA can be expressed as:

\begin{equation*}
\begin{aligned}
\Delta W 
&= \text{Poly}^{2}(B)\text{Poly}^{2}_{\text{H}}(A) \\
&= \sum_{i=1}^{r} \mathbf{b}_{i} \mathbf{a}_{i}^{T} + \sum_{1\le i=j}^{r} h_{ij}(\mathbf{b}_{i}\odot \mathbf{b}_{j}) (\mathbf{a}_{i}^{T}\odot \mathbf{a}_{j}^T) + \sum_{1 \le i < j }^r h_{ij}(\mathbf{b}_{i} \odot \mathbf{b}_{j})(\mathbf{a}_{i}^T \odot \mathbf{a}_{j}^T).
\end{aligned}
\end{equation*}

\subsection{Proof of PERA’s Expressive Power}
In this section, we give the details proof of the expressive power of HiRA in comparison to LoRA. We begin by introducing the Eckart-Young-Mirsky Theorem~\cite{eckart1936approximation}, which provides the optimal low-rank approximation of a matrix. We will refer to this theorem as Lemma 1.
\paragraph{Lemma 1 (Eckart-Young-Mirsky Theorem).}
The best rank-$r$ approximation of a matrix $W$ in the spectral norm is given by the $(r+1)$-th largest singular value, i.e.,
\begin{equation*}
\min_{\mathrm{Rank}(\hat{W}) \le r} \| W - \hat{W} \|_2 = \sigma_{r+1}(W).
\end{equation*}

\paragraph{Theorem 1 (The Expressive Power of PERA).}
Let $\bar{E}$ denote the optimal parameter update. The PERA update is defined as:
\begin{equation*}
\Delta W_{\text{PERA}} = BA + \Delta W_{\text{square}} + \Delta W_{\text{cross}},
\end{equation*}
Where
\begin{itemize}
    \item $\mathrm{Rank}(BA) \le r$,
    \item $\mathrm{Rank}(\Delta W_{\text{square}}) \le r$,
    \item $\mathrm{Rank}(\Delta W_{\text{cross}}) \le C(r,2)$.
\end{itemize}
Then the following bound holds:
\begin{equation*}
\min_{\Delta W \in S_{\text{PERA}}} \| \Delta W - \bar{E} \|_2 \le \sigma_{2r+C(r,2)+1}(\bar{E}),
\end{equation*}
Where
\begin{equation*}
S_{\text{PERA}} = BA + \Delta W_{\text{square}} + \Delta W_{\text{cross}}.
\end{equation*}

\paragraph{Proof.}
Since
\begin{equation*}
\mathrm{rank}(\Delta W_{\text{PERA}}) 
\le \mathrm{rank}(BA) + \mathrm{rank}(\Delta W_{\text{square}}) + \mathrm{rank}(\Delta W_{\text{cross}})
\le 2r + C(r,2)
\end{equation*}
The PERA hypothesis space is contained in the set of matrices with rank at most $2r + C(r,2)$. Applying Lemma 1 yields
\begin{equation*}
\min_{\Delta W \in S_{\text{PERA}}} \| \Delta W - \bar{E} \|_2 \le \sigma_{2r+C(c,2)+1}(\bar{E}).
\end{equation*}

\paragraph{Comparison with LoRA.}
For LoRA,
\begin{equation*}
\mathrm{rank}(\Delta W_{\text{LoRA}}) \le r,
\end{equation*}
Which implies
\begin{equation*}
\min_{\Delta W_{\text{LoRA}}} \| \Delta W - \bar{E} \|_2 = \sigma_{r+1}(\bar{E}).
\end{equation*}
Since singular values are non-increasing
\begin{equation*}
\sigma_{2r + C(r,2) + 1}(\bar{E}) \le \sigma_{r+1}(\bar{E}),
\end{equation*}
Which establishes that PERA admits a strictly larger approximation space in terms of rank-constrained spectral approximation.

\twocolumn

\section{Experimental Details}
\subsection{Datasets}
\label{Appendix: B.1}
\paragraph{Commonsense170K.} This dataset integrates eight widely used benchmarks for commonsense reasoning:
\begin{itemize}
    \item BoolQ~\citep{clark2019boolq}: This dataset consists of 15,942 naturally occurring yes/no questions collected from unprompted and unconstrained settings.
    \item PIQA~\citep{bisk2020piqa}: A multiple-choice dataset focused on physical commonsense reasoning, where each example presents a question with two possible solutions.
    \item SIQA~\citep{sap2019socialiqasiqa}: This dataset contains multiple-choice questions designed to evaluate a model’s ability to reason about the social and pragmatic implications of everyday events. 
    \item HellaSwag~\citep{zellers2019hellaswag}: A commonsense natural language inference benchmark in which models must select the most plausible ending given a contextual prompt.
    \item WinoGrande~\citep{sakaguchi2021winogrande}: A fill-in-the-blank task with two candidate answers, aimed at assessing commonsense reasoning and coreference resolution.
    \item ARC-e and ARC-c~\citep{clark2018thinkarc-e}: The Easy and Challenge subsets of the ARC benchmark, consisting of grade-school-level multiple-choice science questions. Notably, the Challenge set includes questions that are difficult for both retrieval-based and word co-occurrence-based methods.
    \item OBQA~\citep{mihaylov2018canOBQA}: A dataset of elementary-level multiple-choice science questions that require multi-step reasoning, integration of external commonsense knowledge, and rich text comprehension.
\end{itemize}

\paragraph{GLUE.}
The General Language Understanding Evaluation (GLUE) benchmark~\citep{wang2018glue} comprises a collection of datasets designed to evaluate diverse natural language understanding capabilities. These include CoLA~\citep{warstadt2019neural}, SST-2~\citep{socher2013recursive}, MRPC~\citep{dolan2005automatically}, QQP, STS-B~\citep{cer2017semeval}, MNLI~\citep{williams2018broad}, QNLI~\citep{demszky2018transforming}, and RTE~\citep{haim2006second}. In our experiments, we evaluate PERA on six representative datasets from the GLUE benchmark.
For fair comparison, we follow the same dataset splits as in~\citep{wu2024reft, bini2025delora}. Specifically, if a validation set contains more than 2,000 samples, we randomly sample 1,000 examples to form a validation subset and use the remaining samples as the test set. The sizes of the training, validation, and test sets for all datasets are reported in Table~\ref{tab:tab6}.

\begin{table}[htbp]
\centering
\small
\begin{tabular}{lcccc}
\toprule
\textbf{HyperParameters} 
& \multicolumn{2}{c}{\textbf{LLaMA2-7B}} 
& \multicolumn{2}{c}{\textbf{LLaMA3-8B}} \\
\midrule
Rank $r$        & 16 & 32 & 16 & 32 \\
$\alpha$        & 16 & 32 & 16 & 32 \\
Dropout         & \multicolumn{4}{c}{0.05} \\
Optimizer       & \multicolumn{4}{c}{AdamW} \\
LR              & \multicolumn{4}{c}{1e-4} \\
LR Scheduler    & \multicolumn{4}{c}{Linear} \\
Batch Size      & \multicolumn{4}{c}{16} \\
Warmup Steps    & \multicolumn{4}{c}{100} \\
Epochs          & \multicolumn{4}{c}{3} \\
Where           & \multicolumn{4}{c}{Q, K, V, Up, Down} \\
\bottomrule
\end{tabular}
\caption{The hyperparameters for PERA on the commonsense reasoning tasks.}
\label{tab:tab6}
\end{table}

\subsection{Hyperparameters}
\label{Appendix: B.2}

\paragraph{Commonsense Reasoning.} 
Table~\ref{tab:tab7} presents the hyperparameter configurations used when fine-tuning the LLaMA2-7B and LLaMA3-8B models with PERA on the commonsense reasoning. To ensure fair comparisons among LoRA, DoRA, MoRA, and HiRA, we follow the experimental setup in~\citep{huang2025hira} and maintain the same or comparable numbers of trainable parameters.

\begin{table*}
\centering
\small
\begin{tabular}{lcccccccc}
\toprule
\textbf{Splits Sizes}  & \textbf{SST-2} & \textbf{MRPC} & \textbf{CoLA} & \textbf{QNLI}  & \textbf{RTE} & \textbf{STS-B} \\
\midrule
Training Set          & 67K    & 3.7K    & 8.5K    & 105K    & 2.5K  & 5.7K    \\
New Validation Set         & 436    & 204     & 522     & 1K      & 139    & 750     \\
New Test Set           & 436    & 204     & 521     & 4.5K    & 138    & 750     \\
\bottomrule
\end{tabular}
\caption{GLUE dataset sizes, with new validation and test splits following \citep{wu2024reft,bini2025delora} setup}
\label{tab:tab7}
\end{table*}

\begin{table*}[htbp]
\centering
\small
\begin{tabular}{clcccccccc}
\toprule
Model & Hyperparameters  & SST-2 & MRPC & CoLA & QNLI  & RTE  & STS-B \\
\midrule
\multirow{6}{*}{RoBERTa-base}
&Learning Rate    & 1e-4  & 1e-3 & 1e-3 & 3e-4  & 5e-4 & 1e-3 \\
&Batch Size       & 32    & 32   & 32   & 32    & 32   & 32     \\
&Num. Epochs      & 50    & 50   & 40   & 30    & 75   & 80    \\
&Dropout          & 0     & 0.25  & 0    & 0.25  & 0    & 0.25   \\
&Warmup Ratio     & \multicolumn{6}{c}{6e-2} \\
&Maximum Sequence Length &  \multicolumn{6}{c}{512}  \\
\midrule
\multirow{6}{*}{RoBERTa-large}
&Learning Rate    &  5e-4 &1e-4  & 4e-4 &  3e-4 & 5e-4 &  4e-4\\
&Batch Size       & 32    & 32   & 32   & 32    & 32   & 32     \\
&Num. Epochs      &   45  & 35   &  60  &   50  &  75  &  30   \\
&Dropout          &   0.25   &  0.0 & 0.20    & 0.0  &  0.25   & 0.10   \\
&Warmup Ratio     & \multicolumn{6}{c}{6e-2} \\
&Maximum Sequence Length &  \multicolumn{6}{c}{512}  \\
\bottomrule
\end{tabular}
\caption{GLUE benchmark hyperparameters.}
\label{tab:tab8}
\end{table*}

\paragraph{GLUE.}
Following~\citep{bini2025delora}, we use the new validation set to fine-tune the hyperparameters on random seed 42. Then, we select the best hyperparameters to evaluate test performance for seeds 42, 43, 44, 45, 46. For each training run, we use checkpointing to save the best training run and evaluate with that. For all experiments, we use a max sequence length of 512. For all hyperparameters, we run a small grid search. The detailed hyperparameters are reported in Table~\ref{tab:tab8}. 
\section{Detailed Experimental Results}
\subsection{Different Rank Setting}
\label{Appendix: C.1}
Table~\ref{tab:tab9} illustrates the impact of the parameter rank $r$ on model performance across multiple commonsense reasoning tasks when fine-tuning the LLaMA3-8B model.

\subsection{Different Placement in Transformers}
\label{Appendix: C.2}
Table~\ref{tab:tab10} shows the effect of applying PERA to different weight modules within the Transformer architecture. Each module is denoted by its initial letter: (Q)uery, (K)ey, (V)alue, (U)p, and (D)own. We fine-tune the LLaMA3-8B model with the rank set to $16$.

\subsection{Low-Resource Setting}
\label{Appendix: C.3}
Table~\ref{tab:tab11} presents the detailed results under low-resource settings for commonsense reasoning tasks. We fine-tune the LLaMA3-8B model with the rank set to $16$.

\subsection{Different High-Order Components}
\label{Appendix: C.4}
This section presents a detailed empirical analysis of how different high-order components affect LoRA. We evaluate commonsense reasoning tasks using the LLaMA3-8B model with rank fixed at $r=16$. The results are summarized in Table~\ref{tab:tab12}.

Although the square-only variant achieves the highest overall average performance, a finer-grained comparison across individual benchmarks reveals that cross terms consistently yield gains on several tasks. In particular, PERA with cross terms outperforms the square-only variant on SIQA (83.16 vs. 82.34, +0.82), ARC-e (93.73 vs. 92.80, +0.93), and HellaSwag (95.94 vs. 95.83, +0.11). These datasets typically involve multi-hop or compositional reasoning, where inter-component coupling plays a more critical role.

In contrast, on datasets such as ARC-c and PIQA, the square-only formulation performs slightly better. Overall, these results suggest that cross terms provide complementary inter-component interactions that are particularly beneficial for reasoning-intensive benchmarks, while their effectiveness may vary depending on task structure and complexity.

\subsection{Training Loss Analysis}
\begin{figure*}
    \centering
    \includegraphics[width=0.7\linewidth]{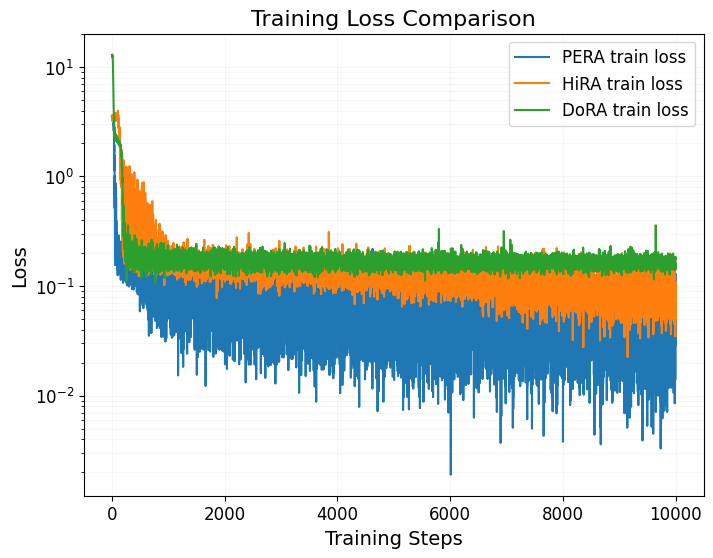}
    \caption{Loss Comparision between Different PEFT Methods}
    \label{fig:fig6}
\end{figure*}
To evaluate the fitting capacity and training efficiency of PERA under a comparable parameter budget, we train all baseline methods—including representative parameter-efficient fine-tuning approaches such as DoRA and HiRA—using the same rank setting ($r=4$) and a unified set of hyperparameters. Figure~\ref{fig:fig6} presents a comparison of the training loss trajectories across different methods.

At the early stage of training, all three methods exhibit rapid convergence; however, PERA shows a noticeably steeper decline, indicating stronger initial fitting dynamics. As training progresses into a more stable phase, PERA consistently achieves lower training loss than both DoRA and HiRA, maintaining the best convergence bound throughout optimization.

Ultimately, PERA reaches a final training loss of 0.0425, whereas DoRA and HiRA obtain 0.1595 and 0.0915, respectively. These results suggest that, under the same rank constraint, PERA’s higher-order nonlinear structure enables it to capture more complex feature mappings than conventional linear low-rank adaptation methods, thereby achieving superior task performance with the same parameter efficiency.

\begin{table*}[h]
\centering
\small
\begin{tabular}{lcccccccccc}
\toprule
\textbf{Rank} & \textbf{\#Param (\%)} & \textbf{BoolQ} & \textbf{PIQA} & \textbf{SIQA} & \textbf{ARC-c} & \textbf{ARC-e} & \textbf{OBQA} & \textbf{HellaS} & \textbf{WinoG} & \textbf{Avg.} \\
\midrule
2 & 0.0440 & 72.39 & 89.39 & 83.32 & 83.45 & 93.43 & 88.00 & 96.14 & 89.19 & 86.91 \\
4 & 0.0880 & 74.25 & 89.93 & 81.88 & 83.28 & 93.56 & 87.60 & 96.48 & 89.11 & 87.01 \\
8 & 0.1760 & 75.72 & 88.68 & 81.78 & 83.53 & 93.64 & 88.00 & 96.15 & 88.95 & 87.05 \\
12 & 0.2638 & 75.11 & 87.49 & 82.60 & 83.79 & 93.56 & 88.40 & 95.87 & 89.74 & 87.07 \\
16 & 0.3515 & 76.15 & 88.57 & 83.16 & 83.62 & 93.73 & 88.60 & 95.94 & 89.27 & 87.38 \\
20 & 0.4391 & 74.62 & 89.83 & 81.93 & 83.28 & 93.86 & 90.20 & 95.77 & 88.56 & 87.25 \\
24 & 0.5266 & 76.06 & 89.66 & 82.34 & 84.39 & 93.65 & 89.00 & 95.71 & 88.71 & 87.44 \\
32 & 0.7012 & 75.54 & 89.83 & 81.58 & 83.53 & 93.56 & 88.40 & 95.61 & 88.95 & 87.12 \\
\bottomrule
\end{tabular}
\caption{Accuracy (\%) comparison with the rank $r$ increases on LLaMA3-8B model.}
\label{tab:tab9}
\end{table*}

\begin{table*}[htbp]
\centering
\small
\begin{tabular}{lcccccccccc}
\toprule
\textbf{Placement} & \textbf{\#Params (\%)} & \textbf{BoolQ} & \textbf{PIQA} & \textbf{SIQA} & \textbf{ARC-c} & \textbf{ARC-e} & \textbf{OBQA} & \textbf{HellaS} & \textbf{WinoG} & \textbf{Avg.} \\
\midrule
QKV, UD & 0.3513& \textbf{76.15} & 88.57 & \textbf{83.16} & \textbf{83.62} & 93.73 & 88.60 & 95.94 & \textbf{89.27} & \textbf{87.38} \\
QKV & 0.1176 & 74.71 & \textbf{89.39} & 82.50 & \textbf{83.62} & \textbf{93.81} & 87.60 & 95.92 & 88.63 & 87.02 \\
UD & 0.2345 & 72.66 & 89.61 & 82.45 & 83.53 & 93.39 & \textbf{89.00} & \textbf{96.25} & 88.40 & 86.91 \\
QK & 0.0849 &71.96 & 88.30 & 80.96 & 80.12 & 93.01 & 86.80 & 95.01 & 85.87 & 85.25 \\
QV & 0.0849 & 74.80 & 89.12 & 81.73 & 83.19 & 93.01 & 87.00 & 96.03 & 88.71 & 86.70 \\
Q & 0.0522 &71.80 & 88.63 & 80.71 & 79.78 & 92.26 & 84.60 & 94.99 & 85.87 & 84.83 \\
K & 0.0326 &70.86 & 87.65 & 78.25 & 78.67 & 91.96 & 82.80 & 93.84 & 84.21 & 83.53 \\
V & 0.0326 &71.93 & 88.63 & 82.60 & 82.42 & 93.52 & 87.60 & 95.69 & 87.13 & 86.19 \\
\bottomrule
\end{tabular}
\caption{Accuracy (\%) comparison of different placement on commonsense reasoning tasks with LLaMA3-8B model.}
\label{tab:tab10}
\end{table*}

\begin{table*}[h]
\centering
\small
\begin{tabular}{c|ccccccccc}
\toprule
\textbf{Pruning Ratio} & \textbf{BoolQ} & \textbf{PIQA} & \textbf{SIQA} & \textbf{ARC-c} & \textbf{ARC-e} & \textbf{OBQA} & \textbf{HellaS} & \textbf{WinoG} & \textbf{Average}  \\
\midrule
0.1 & 68.78 & 86.67 & 77.38 & 80.46 & 92.26 & 82.60 & 92.03 & 84.37 & 83.07 \\
0.2 & 68.84 & 84.93 & 77.79 & 79.44 & 91.75 & 81.60 & 91.04 & 82.40 & 82.22 \\
0.5 & 73.52 & 88.41 & 80.91 & 82.94 & 93.22 & 85.60 & 95.30 & 89.19 & 86.14 \\
0.8 & 75.60 & 89.06 & 82.04 & 84.56 & 93.77 & 87.60 & 96.19 & 88.32 & 87.14 \\
1.0 & 76.15 & 88.57 & 83.16 & 83.62 & 93.73 & 88.60 & 95.94 & 89.27 & 87.38 \\
\bottomrule
\end{tabular}
\caption{Accuracy (\%) comparision of different training samples pruning ratio fine-tuning LLaMA3-8B model on the commonsense reasoning tasks.}
\label{tab:tab11}
\end{table*}

\begin{table*}[h]
\centering
\small
\begin{tabular}{l|ccccccccc}
\toprule
\textbf{Component} & \textbf{BoolQ} & \textbf{PIQA} & \textbf{SIQA} & \textbf{ARC-c} & \textbf{ARC-e} & \textbf{OBQA} & \textbf{HellaS} & \textbf{WinoG} & \textbf{Average}  \\
\midrule
LoRA & 72.30 & 86.70 & 79.30 & 75.70 & 87.70 & 82.80 & 93.50 & 84.80 & 82.80 \\
LoRA + cross features & 75.41 & 88.52 & 81.78 & 82.76 & 93.05 & 88.80 & 95.75 & 88.55 & 86.83 \\
LoRA + square features & 76.17 & 89.28 & 82.34 & 84.72 & 92.80 & 89.20 & 95.83 & 89.50 & 87.48 \\
PERA & 76.15 & 88.57 & 83.16 & 83.62 & 93.73 & 88.60 & 95.94 & 89.27 & 87.38 \\
\bottomrule
\end{tabular}
\caption{Accuracy (\%) comparision of different high-order components from LoRA with LLaMA3-8B model on the commonsense reasoning tasks.}
\label{tab:tab12}
\end{table*}

\end{document}